\documentclass[10pt,twocolumn,letterpaper]{article}

\usepackage{cvpr}              
\usepackage{makecell}
\usepackage{subcaption}
\usepackage{graphicx}
\usepackage{listings}
\usepackage{pgfplots}
\usepackage{framed}
\usepackage{multicol}
\usepackage[utf8]{inputenc}
\usepackage{amsmath}
\usepackage[linesnumbered,ruled,vlined]{algorithm2e}
\newcommand{\pipeline}{{Scene Copilot}}
\definecolor{cvprblue}{rgb}{0.21,0.49,0.74}
\usepackage[pagebackref,breaklinks,colorlinks,allcolors=cvprblue]{hyperref}

\def\paperID{8874} 
\def\confName{CVPR}
\def\confYear{2025}

\title{Scene Co-pilot: Procedural Text to Video Generation with Human in the Loop}

\author{
\small 
\begin{tabular}{cccc}
Zhaofang Qian & Abolfazl Sharifi & Tucker Carroll & Ser-Nam Lim \\
University of Central Florida & University of Kashan & University of Central Florida & University of Central Florida \\
{\tt\scriptsize zh103512@ucf.edu} & {\tt\scriptsize sharifi.abolfazl@grad.kashanu.ac.ir} & {\tt\scriptsize tu512807@ucf.edu} & {\tt\scriptsize sernam@ucf.edu} \\
\end{tabular}
}

\begin{document}
\maketitle
\begin{abstract}
Video generation has achieved impressive quality, but it still suffers from artifacts such as temporal inconsistency and violation of physical laws. Leveraging 3D scenes can fundamentally resolve these issues by providing precise control over scene entities. To facilitate the easy generation of diverse photorealistic scenes, we propose \pipeline{}, a framework combining large language models (LLMs) with a procedural 3D scene generator. Specifically, \pipeline{} consists of Scene Codex, BlenderGPT, and Human in the loop. Scene Codex is designed to translate textual user input into commands understandable by the 3D scene generator. BlenderGPT provides users with an intuitive and direct way to precisely control the generated 3D scene and the final output video. Furthermore, users can utilize Blender UI to receive instant visual feedback. Additionally, we have curated a procedural dataset of objects in code format to further enhance our system's capabilities. Each component works seamlessly together to support users in generating desired 3D scenes. Extensive experiments demonstrate the capability of our framework in customizing 3D scenes and video generation. For the best visualization, please visit our project page \url{https://abolfazl-sh.github.io/Scene_co-pilot_site/}.
\end{abstract}
\vspace{-0.25cm}
    
\section{Introduction}
\label{sec:intro}
Video generation has emerged as a vibrant and actively investigated area of research. Current state-of-the-art models such as Open AI's Sora\,\cite{sora}, Kling\,\cite{klingai}, and StreamingT2V\,\cite{streamingt2v} can produce movie-quality videos of several minutes in length. However, both industrial and research-level models suffer from numerous limitations ranging from unrealistic and incoherent motion, intermittent object disappearances, to incorrect interactions between multiple objects\,\cite{sun2024sora}. Two particularly noticeable and counterintuitive issues are violations of physical laws and temporal inconsistencies.  

To address these challenges, one promising direction is to generate videos based on a preexisting scene. 
By creating 3D entities within the scene, the aforementioned limitations can be automatically avoided, and the generated videos could theoretically be infinitely long.
Embodied AI and applications in autonomous driving have motivated the community to create numerous scene datasets such as WayveScenes101\,\cite{zürn2024wayvescenes101datasetbenchmarknovel}, nuScenes\,\cite{caesar2020nuscenesmultimodaldatasetautonomous}, and EmbodiedScan\,\cite{embodiedscan}. However, the creation and collection of these datasets can be a time-intensive process. Infinigen, a photorealistic procedural scene generator based on Blender\,\cite{blender}, has demonstrated the potential to alleviate such laborious work\,\cite{infinigen2023infinite, infinigen2024indoors}. 

Meanwhile, large language models (LLMs) have demonstrated outstanding ability to efficiently assist with various tasks across different domains\,\cite{wermelinger2023using, codex,lu2024multimodal}. Inspired by this, we've enhanced Infinigen's capabilities by selectively incorporating LLMs. Additionally, we've expanded upon its supporting assets by curating a dataset of procedural objects. 
Our dataset consists of over 300 node-based assets in code format, enabling LLMs to creatively combine assets according to user requests. We also include a backup modular text-to-3D model to generate the required elements when using geometry nodes. 

While Infinigen is powerful, it remains esoteric because of its prerequisite in comprehensive understanding of generation rules and mathematical algorithms. Procedural generation with geometry nodes in Blender would remain niche without an intuitive user interface. Utilizing our dataset, we propose a framework named \pipeline{} to integrate Infinigen with LLMs, making the pipeline more controllable and user-friendly. Leveraging existing LLMs also renders our system training-free and updatable as more powerful models become available (Sec. \ref{sec:addmore}). 

Specifically, given a user's textual input, the first LLM interprets the prompt, incorporating additional related contexts from the Infinigen codebase, and generates a base scene Python command. The initial base scene with a simple and crude setting is generated, allowing users to control the details and swiftly update the object settings in the scene with the help of another LLM. Intermediate users can directly interact with the scene using Blender's graphical user interface (GUI), while beginners can also utilize advanced Blender features with the assistance of the LLM. After updating the base scene, Infinigen continues to add more details. Users can review the finished scene and make further updates before the rendering process starts. As such, with the help of LLMs, general users can benefit from the advanced features in Infinigen and Blender as these two powerful engines are updated.

Our contributions can be summarized as follows:
\begin{itemize}[leftmargin=10pt,noitemsep]
\item We've curated a procedural asset dataset in code format.
\item We've proposed a training-free framework \pipeline{} and demonstrated its controllability on generated scenes and videos. 
\end{itemize}
\section{Related Work}
\label{sec:related_work}

\subsection{Synthesis Scene Dataset}
Virtual 3D scenes have been widely deployed in fields such as simulation, computer vision, and interactive systems\,\cite{vidanapathirana2021plan2scene, mao2021one, xu2024sketch2scene, yang2024physcene}. However, real-world scene data is limited due to the laborious and time-consuming nature of data collection. To address the scarcity of real-world data, researchers have turned to synthetic data generation\,\cite{butler2012naturalistic, deitke2022️, fabbri2021motsynth}. Hypersim\,\cite{roberts2021hypersim} is an indoor synthetic scene dataset containing objects such as chairs, TVs, tables, and lamps. In contrast, GOS\,\cite{xie2022gos} offers outdoor models suitable for tasks that rely on outdoor scene representations.
Infinigen\,\cite{infinigen2023infinite} is a framework designed to procedurally generate photorealistic 3D scenes within Blender, with the generation process guided by Python APIs. The framework includes a foundational dataset, and due to the randomization of parameters in the generation process, each generated scene is unique. 
\subsection{Procedural Generation}
Procedural Content Generation has been extensively studied and applied in industrial production\,\cite{shaker2016procedural, liu2021deep, risi2020increasing, gasch2022procedural}. Recently, Blender introduced geometry nodes to provide an intuitive UI for artists. Procedural generation using geometry and material nodes allows for automatic manipulations of models\,\cite{hu2022inverse, shi2020match, zhang2019procedural, gasch2022procedural}. The authors in \cite{hu2023generating} propose a framework that creates Geometry-Graph materials based on text conditions and images. Text conditions are converted into embedded vectors, which are then mapped to corresponding images. Subsequently, procedural materials are generated from these images. The resulting materials are structured as node graphs, facilitating user-driven modifications. Additionally, a procedural material dataset is introduced to enhance the quality of generated materials.

\subsection{Large Language models}

Large Language Models such as GPT-4o\,\cite{gpt4o} and Claude-3.5-Sonnet\,\cite{claude-3-5} have demonstrated exceptional capabilities in coding, reasoning, and language comprehension.
Their potential in code understanding and generation can be leveraged for procedural 3D modeling, where user prompts are converted into corresponding code by the model. LLplace\,\cite{yang2024regionplc} utilizes LLMs to transform user prompts into corresponding code to create 3D scenes, allowing users to precisely specify the location of target objects.

Fine-tuning LLMs is a prominent approach to improve the quality of generated 3D models\,\cite{brown2020language, liu2021makes, min2022rethinking}. Incorporating contextual information into the 3D model generation process enables LLMs to operate with greater precision. For instance, Liu et al.\,\cite{liu2024uni3d} focus on a Multi-modal Large Language Model (MLLM) for generating 3D models from user prompts, enhancing the user's input description prior to model generation. The integration of extensive linguistic knowledge facilitates the analysis of the semantics within user prompts, enabling a more accurate understanding of user intentions\,\cite{ding2023pla, yang2024regionplc, lu2023ovir, cao2024coda}. 
Fang et al.\,\cite{fang2025chat} employ LLMs to enhance the flexibility of scene editing for users. In their method, the 3D model is converted into 2D space using a Hash-Atlas network. Editing models in 2D space simplifies the complexity of fusion design. 
Additionally, few-shot learning approaches have become a popular strategy for enhancing LLM-based methods\,\cite{tang2024minigpt,zhu2024llafs,ma2024llms,fu2024scene}. These approaches provide structured guidance for generating 3D models from a limited number of examples. This guidance reduces processing time while improving output model quality.

\subsection{Text-to-Scene}


Recent progress in Text-to-image models have promoted more precise text-to-scene generation\,\cite{jain2022zero,khalid2023clip,lee2022understanding}. In \cite{zhang2024text2nerf}, the authors propose a method for converting user prompts into photorealistic scenes by transforming prompts into image priors from different angles, which are then to NeRF model. SceneScape\,\cite{fridman2024scenescape}, converts a given prompt into a series of images with overlapping and shared segments, which are subsequently used to generate a long-term video that depicts a walkthrough of the scene. Additionally, several methods rely on large image datasets \cite{jain2021putting,yu2021pixelnerf,xu2022sinnerf}. These methods map text to corresponding images within the dataset and synthesize scenes based on the matched visual content.

Agent-based approaches have also been explored\,\cite{hu2024scenecraft,sun20243dgptprocedural3dmodeling, zhou2024scenexproceduralcontrollablelargescalescene}. In \cite{sun20243dgptprocedural3dmodeling}, the authors present a text-to-3D model platform named 3D-GPT. The proposed method employs a multi-agent system for reasoning, planning, and tool utilization to convert user prompts into 3D models. \cite{zhou2024scenexproceduralcontrollablelargescalescene} presents a similar approach based on multi-agent systems to manage scene generation. In this framework, a task-planning agent assigns various tasks to subordinate agents to accomplish the primary objective.



\section{Dataset}

\label{sec:dataset}

Collecting procedural objects is a time-consuming and tedious task as assets are limited in both quantity and quality. In this work, we aim to address this gap by compiling a procedural dataset to support LLM-driven code generation approaches. To achieve this, we systematically gathered Blender-based procedural geometry and materials assets from the Internet. After constructing the dataset, we utilized the Node Transpiler from Infinigen and Node2Python\,\cite{node2python} to convert assets and materials into Python code. This conversion enables automated modification of asset features, facilitating the generation of new assets.

\subsection{Dataset Details}
Our dataset consists of 323 procedural assets and materials, covering a wide range of categories to provide a comprehensive resource (Table\,\ref{table:dataset_details}).

\begin{table}[h]
    \centering
    \begin{tabular}{|c|c|} \hline  
         Assets Category& Num. Generator\\ \hline  
         Indoors& 29\\ \hline  
         Outdoors& 8\\ \hline  
 Terrain&6\\ \hline  
 Rocks&7\\ \hline  
         Plants& 17\\ \hline  
         Trees& 10\\ \hline  
         Weather& 7\\ \hline  
         Foods& 9\\ \hline  
         Scattering& 24\\ \hline  
         Materials& 206\\ \hline  
 Total&323\\ \hline 
    \end{tabular}
    \caption{Dataset Details: All components are procedural and have been converted to code, allowing for automated modifications.}
    \label{table:dataset_details}
\end{table}

\begin{figure}
    \centering
    \includegraphics[width=1\linewidth]{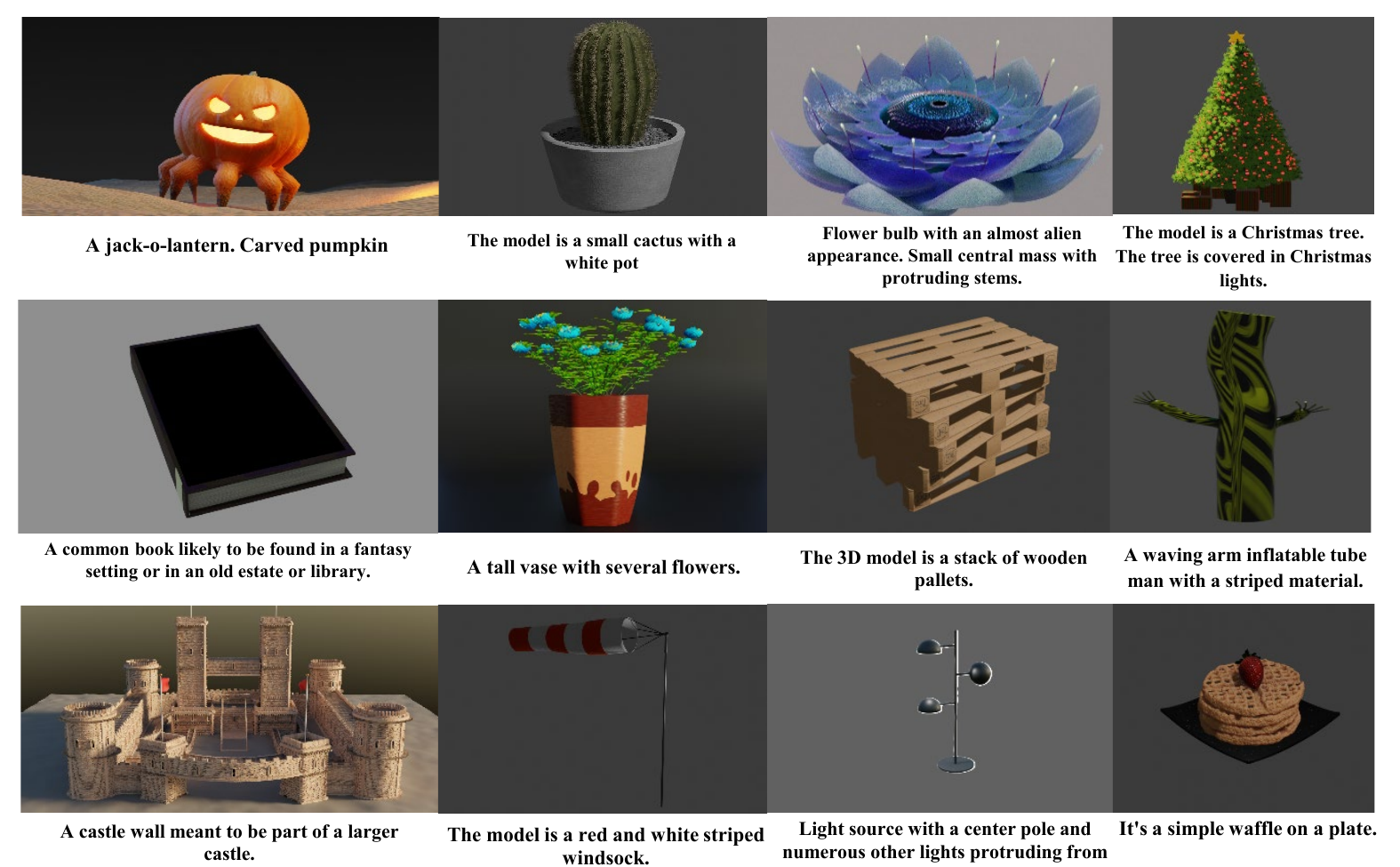}
    \caption{Dataset Samples: A subset of assets from our dataset, accompanied by corresponding descriptions. All assets are fully procedural, meaning their components can be modified either manually or automatically.}
    \label{fig:assets_desc}
\end{figure}

A selection of sample objects is shown in Figure~\ref{fig:assets_desc}. 
In addition to utilizing our primary dataset, we integrated a supplementary dataset to generate assets exhibiting minimal similarity to existing assets within our collection. To achieve this, we employed Shap-E\,\cite{jun2023shape}, a text-to-3D diffusion-based model, to create a diverse array of objects. Shap-E is trained on Neural Radiance Fields (NeRF) and further refined through Signed Distance Fields (SDF) rendering, enhancing the output quality. Notably, Shap-E is a lightweight yet precise model, effectively translating user prompts into detailed 3D objects.
Asset generation was guided by a probability function derived from a Gaussian distribution, which determines whether to use LLMs with our procedural dataset or Shap-E for object creation. Objects generated by Shap-E can also be used in tandem with our procedural dataset or primitive objects (Section \ref{sec:modularity}).

\begin{figure}[h]
    \centering
    \includegraphics[width=1\linewidth]{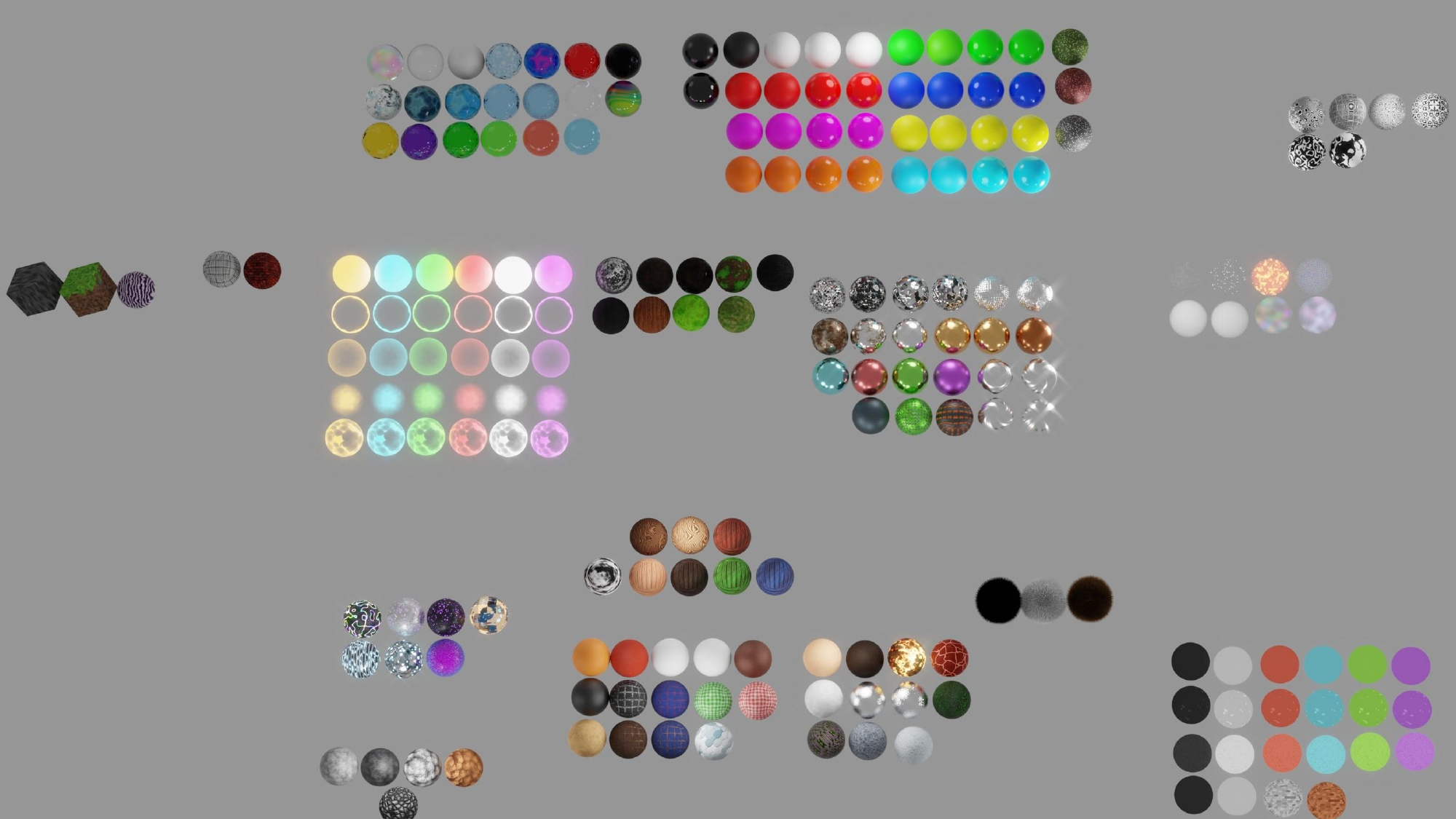}
    \caption{Materials in our dataset can serve as textures, and they can be automatically integrated with various assets, allowing for seamless combination and enhancing visual coherence.}
    \label{fig:materials}
\end{figure}

We have collected 206 materials to facilitate the generation of assets with diverse textures, partially shown in Figure\,\ref{fig:materials}. The integration of these materials with generated assets can be performed either manually or automatically. To streamline the combination process and enable automatic modification of material features, we maintain all materials in code format. 

\begin{figure}[h]
    \centering
    \includegraphics[width=1\linewidth]{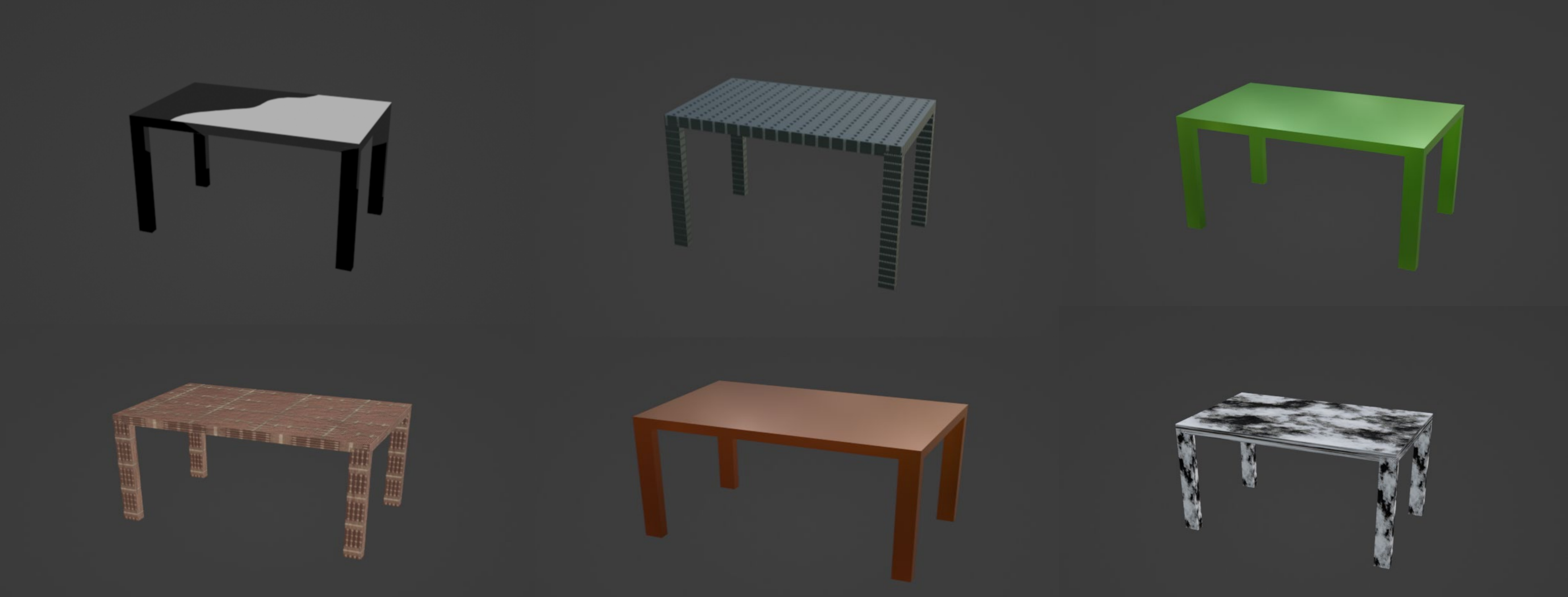}
    \caption{Various types of materials are employed as textures for a table asset,  facilitating the effective fulfillment of user prompts. These materials enable seamless integration and enhance the versatility of the assets to meet diverse user requirements.}
    \label{fig:tables}
\end{figure}

\begin{figure}[h]
    \centering
    \includegraphics[width=1\linewidth]{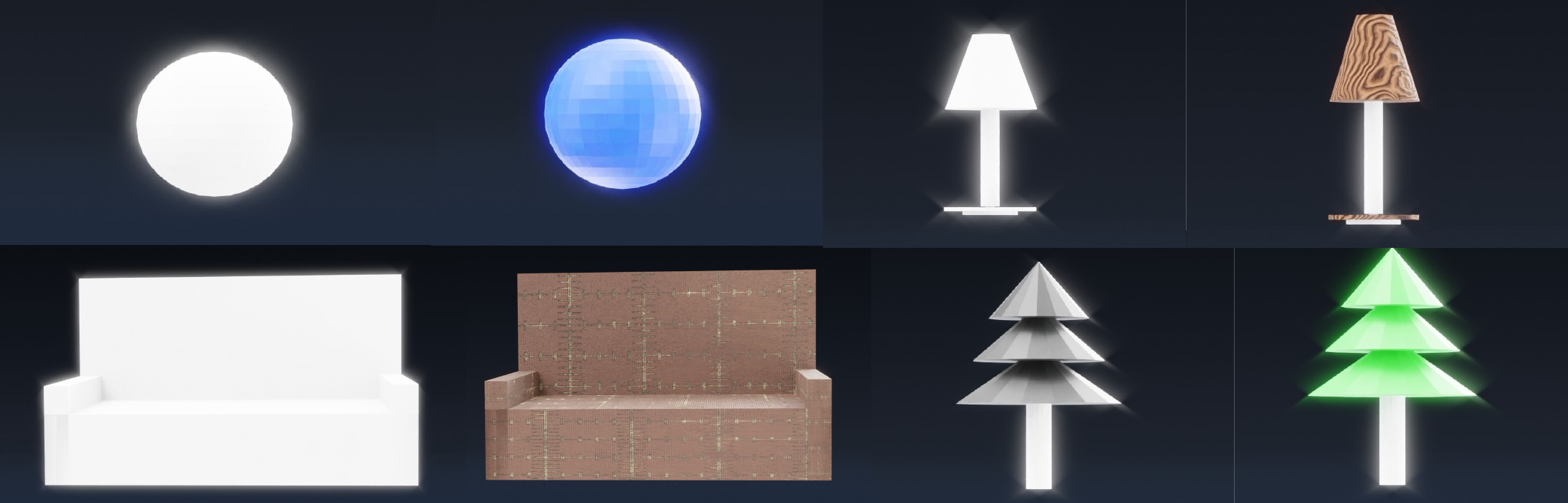}
    \caption{Materials are combined with assets automatically, with selection guided either by user prompts or by identifying materials relevant to each specific asset. Additionally, users have the option to manually apply materials to assets according to their individual requirements.}
    \label{fig:assets_materials}
\end{figure}

As shown in Figure~\,\ref{fig:tables}, our procedural assets can be generated with a variety of material types, enhancing the diversity available for user creation. The integration of materials with procedural assets is demonstrated in Figure\,\ref{fig:assets_materials}. Material selection for each asset can be determined either by user prompts or by analyzing the conceptual attributes of the asset. For instance, the leaves of trees can be rendered in green, yellow, or orange, reflecting natural variations.

\subsection{Modular Objects}
\label{sec:modularity}

\begin{figure}[h]
    \centering

    \begin{subfigure}{0.23\textwidth}
        \centering
        \includegraphics[height=4.2cm, width=0.6\linewidth]{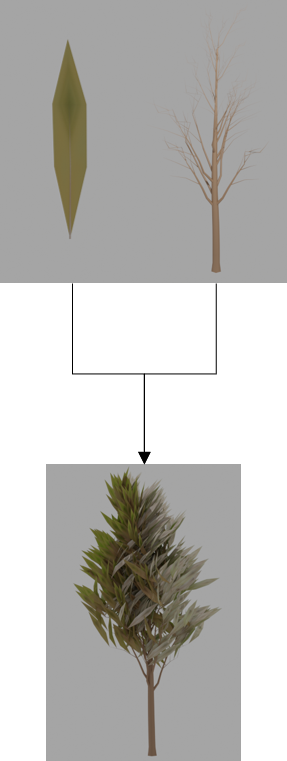}
        \caption{A leaf from our procedural dataset (top left) combined with a tree with no leaves, also from our procedural dataset (top right) to create a fully procedural tree with leaves (bottom).}
        \label{fig:modular_tree}
    \end{subfigure}
    \hfill
    \begin{subfigure}{0.23\textwidth}
        \centering
        \includegraphics[width=\linewidth]{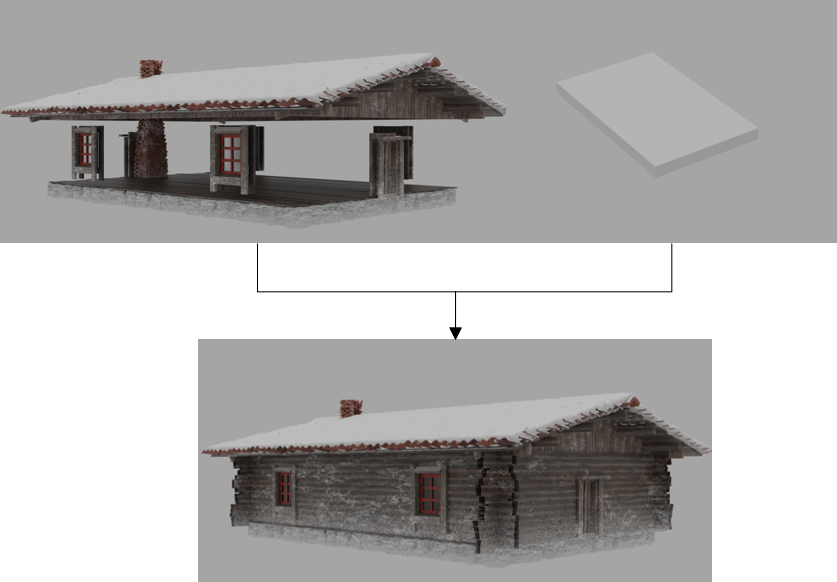}
        \caption{A cottage generator from our procedural dataset (top left) combined with a primitive user-created cube mesh (top right) resulting in a cottage made of logs matching the shape specified by the primitive mesh object (bottom).}
        \label{fig:modular_cottage}
    \end{subfigure}

    \vspace{0.1cm}

    \begin{subfigure}{0.23\textwidth}
        \centering
        \includegraphics[width=\linewidth]{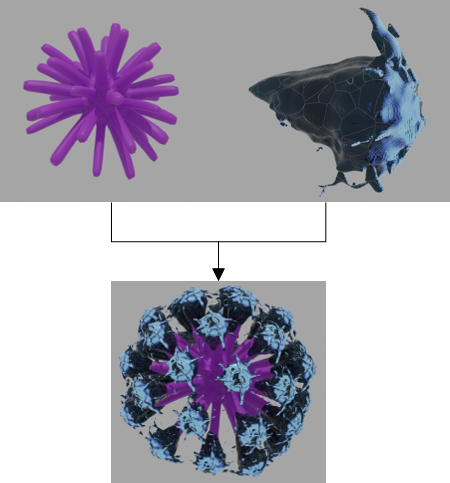}
        \caption{The "FlowerBulb" object from our procedural dataset (top left) and an object generated by Shap-E with the prompt "alien fruit which would be found growing from an alien plant" (top right) combined to create an alien flower with custom fruit (bottom).}
        \label{fig:modular_alienflower}
    \end{subfigure}
    \hfill
    \begin{subfigure}{0.23\textwidth}
        \centering
        \includegraphics[width=\linewidth]{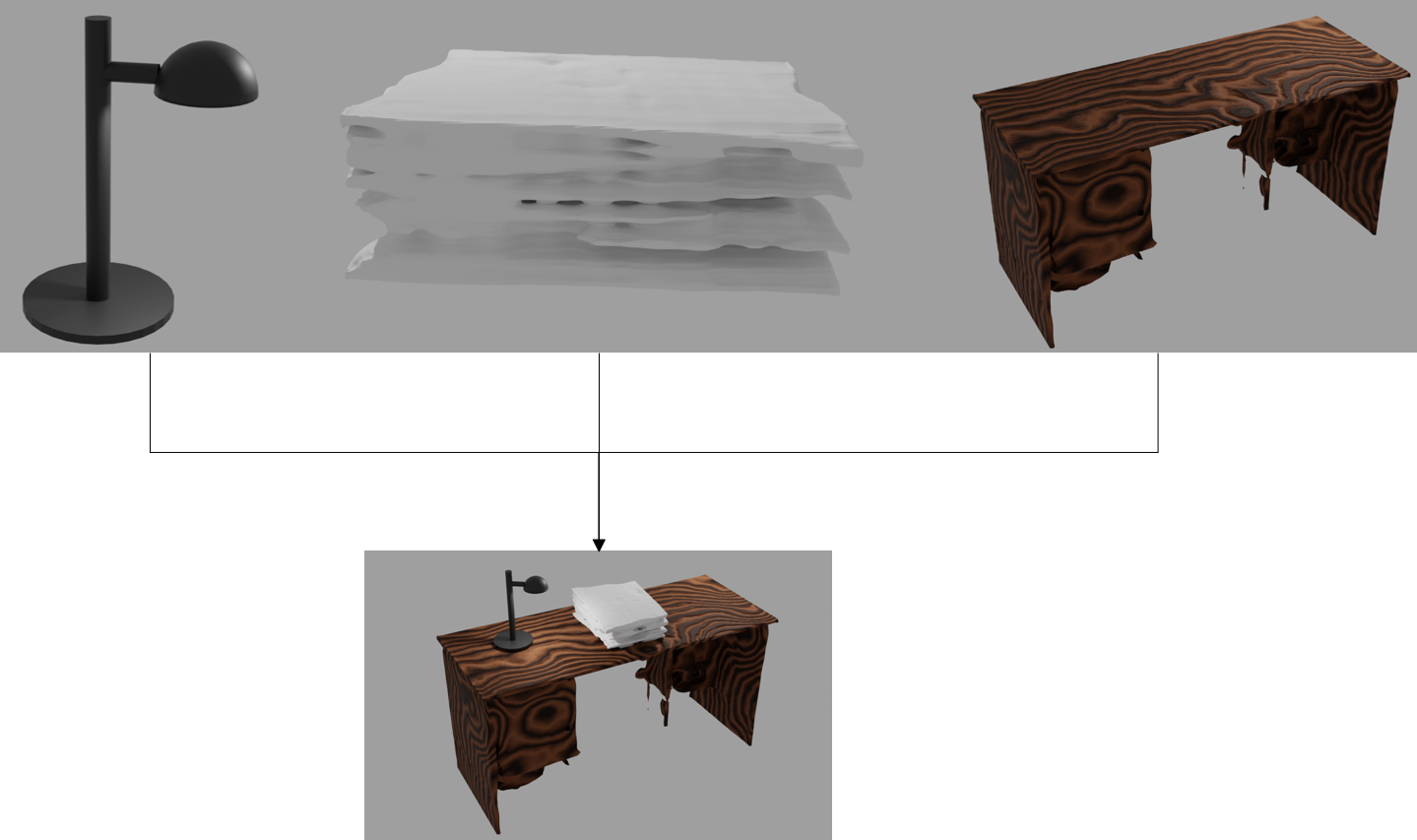}
        \caption{A lamp from our procedural dataset (top left), a ``stack of papers" generated with Shap-E (top middle), and a ``desk" generated with Shap-E (top right) combined to create a more complicated and nuanced object (bottom).}
        \label{fig:modular_desk}
    \end{subfigure}

    \caption{Combining procedural assets and generative models to create diverse, modular objects.}
    \vspace{-0.5cm} 
    
    \label{fig:combining_assets}
    
\end{figure}

Combining Shap-E with our dataset enables the generation of more complex assets or entire scenes. For example, in Figure\,\ref{fig:combining_assets}, Shap-E was used to generate the desk and the stack of papers, while our procedural dataset was used to generate the lamp. Objects themselves can also be modular, allowing them to be combined to inherit or exert traits of other objects, as shown in Figure\,\ref{fig:combining_assets}.

\subsection{Accessibility}

An essential characteristic of any dataset is its accessibility and usability for the broader research community. In this context, the dataset presented in our work represents a significant contribution to the field and constitutes one of the key innovations of this work. The dataset has been meticulously curated to support further research, development, and experimentation. 

This comprehensive dataset includes procedural assets, the corresponding source code, direct links to the assets, and detailed metadata describing each component. The procedural assets are organized in a manner that allows for easy integration and use in related studies, while the descriptions provide contextual information to assist in the understanding and application of each asset. The inclusion of code further enables reproducibility and the validation of results, a core principle in contemporary research.

The dataset is published under the Creative Commons Attribution 4.0 International (CC-BY 4.0) license, ensuring that it can be freely accessed, shared, and adapted, provided proper attribution is given. By making this dataset openly available, we aim to foster collaboration, support future research, and enhance the transparency and reproducibility of experimental work within the community.

\subsection{Expanding upon the Dataset}
\label{sec:addmore}
New procedural assets are released for Blender every single day. As such, the number of procedural assets available is infinitely expanding, the vast majority of which are locked behind paywalls or subscriptions. Thus, in order to keep our dataset free and open-source, we have included scripts that can automatically integrate users' objects into our procedural dataset.

\section{Scene Copilot}
\label{sec:pipeline}
\begin{figure*}[t!]
\centering 
\includegraphics[width=\textwidth]{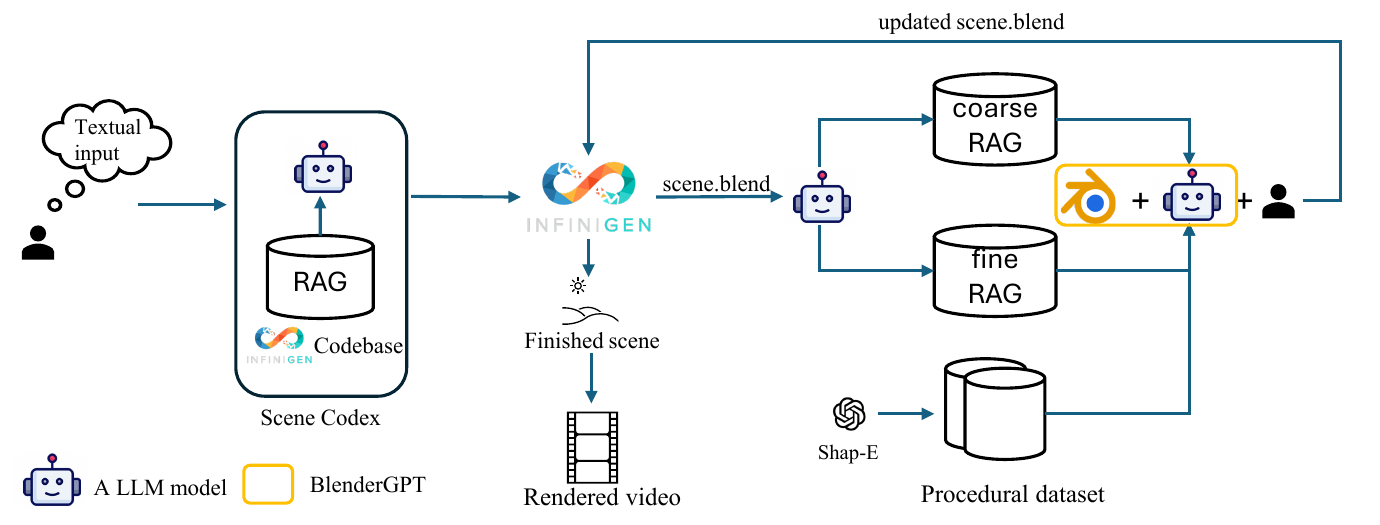}
\vspace{-0.5cm}
\caption{Overview of \pipeline{} with procedural dataset. Starting with a user textual input, Scene Codex combines an LLM with an RAG database of the Infinigen code and generates an Infinigen executable Python command. Infinigen initially creates a coarse scene, which is converted into textual format so that LLMs can comprehend objects and metadata in the scene. Such scene file is condensed into a coarse RAG database. BlenderGPT, incorporating Blender and LLM, utilizes this database to edit and modify the 3D contents in the scene with user involved through either textual or visual interaction. The updated coarse scene is fed back to Infinigen to create a fine scene. Similar to coarse scene, it is condensed into a fine RAG database, and BlenderGPT collaborates both databases while editing the final scene. Meanwhile, The procedural dataset will provide the requested procedural asset code. The finished scene is then rendered by Inifnigen and outputs the requested video.}
\vspace{-0.3cm}
\label{fig:pipeline}
\end{figure*}

To better control the generation of scenes, we combine Infinigen with LLM to generate the necessary commands based on user requests (Sec \ref{sec:codex}). We then utilize another LLM with chain-of-thought to assist users when editing the scene in Blender (Sec \ref{sec:blendergpt}). The full pipeline of \pipeline{} is shown in Figure~\ref{fig:pipeline}.

\subsection{Scene Codex}
\label{sec:codex}
 Procedural generation is powerful yet complex as countless rules and mathematical formulas are involved in each object generator. There are more than 1,000 human-interpretable parameters in Infinigen and even more for low-level external parameters\,\cite{infinigen2023infinite}. This complexity hinders general designers from utilizing Infinigen to swiftly generate 3D scenes or assets. However, LLMs have shown remarkable performance in interpreting large codebase and generating high-quality code\,\cite{llmcode1,codellama,li2024looglelongcontextlanguagemodels}. 
 
Similar to Codex\,\cite{codex}, Scene Codex is designed to convert user intentions into Infinigen commands to create the base scene. To improve the accuracy and correctness of the generated commands, we power our LLM with Retrieval-augmented generation (RAG) \cite{rag} using the Infinigen codebase, given its effectiveness in LLMs answer quality \,\cite{rag_survey}. Guided by few-shot \emph{description-command} example pairs, our Scene Codex converts the user's textual input directly into an Infinigen executable Python command. We applied Claude-3.5-Sonnet as our LLM model and FAISS\,\cite{faiss} as the RAG database. See Appendix for more details on the system prompt and few-shot examples of Scene Codex.\\

\subsection{BlenderGPT}
\label{sec:blendergpt}
As shown in Figure~\ref{fig:pipeline}, Infinigen first generates a coarse scene, and users can edit and update details with our adapted BlenderGPT\footnote{\url{https://github.com/gd3kr/BlenderGPT}}, a Blender add-on with LLMs to directly convert text input to Blender code. We follow the idea of chain-of-thought \cite{cot}, such that BlenderGPT converts the user's prompt into initial steps, reflects on them, and forms the final procedures with objects related to the task. Subsequently, another LLM generates the executable \texttt{bpy} code closely following the procedure, consulting with the textual scene description with the coarse and fine RAG database detailed below. The edited scene is fed back to Infinigen to generate a refined scene with more objects and details. BlenderGPT repeats the previous process to allow users to make a final adjustment. The procedural dataset is integrated with BlenderGPT at both stages to provide the required code of procedural geometry or material nodes. After the user's final confirmation, Infinigen populates the final blender scene and outputs the rendered video. \\

\begin{figure}
    \centering
    \includegraphics[width=0.8\linewidth]{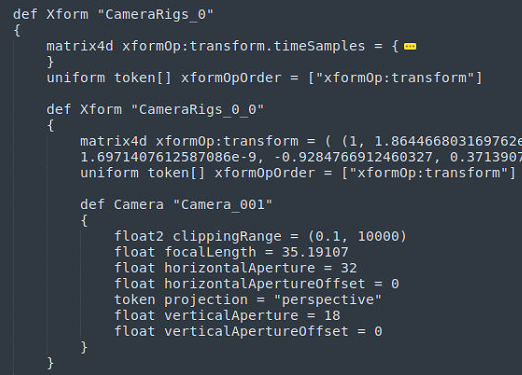}
    \caption{Code segment from converted .usda file. It shows a snippet of camera configurations, including the camera's $ \mathrm{SE}(3) $(xformOp:transform), focal length, and aperture. All these parameters are also available in Blender's GUI. }
    \label{fig:codesnippet}
\end{figure}
\noindent\textbf{Preprocessing Blender Scene} BlenderGPT combines LLMs with Blender so that it can be controlled by natural languages and support multi-round conversations in direct and simple English. However, it defaults from a blank Blender project and lack the ability to understand any imported 3D scenes. Therefore, it is crucial to introduce the base scene to the BlenderGPT. As LLMs can only understand textual input, we convert Blender scenes into \texttt{.usda} files that describe each scene in textual format. Figure~\ref{fig:codesnippet} shows an example of a camera code snippet in a \texttt{.usda} file, which includes metadata ranging from the scene terrain to object animation information. 
After acquiring the raw \texttt{.usda} file that describes the whole scene, we utilize a lightweight LLM (GPT-4o mini\,\cite{gpt4omini}) to further convert and clean the textual contents into a dictionary format and replace the numerical data with \textbf{NUM} placeholder to concise the whole context, such as:
\begin{center}
\begin{lstlisting}
     matrix4d xformOp:transform = {...}
  -> matrix4d xformOp:transform = NUM
\end{lstlisting}
\end{center}

\noindent Such condensation helps LLMs focus on the interpretable parameters instead of being overwhelmed by numerical information. The cleaned scene file is then segmented into non-overlapping chunks and stored as a RAG database. Both the coarse and fine RAG databases are connected to BlenderGPT to help create and interact with the scene objects in later stages. See Appendix for more details on the prompt initialization of BlenderGPT.\\
\begin{figure}
    \centering
    \includegraphics[width=0.8\linewidth]{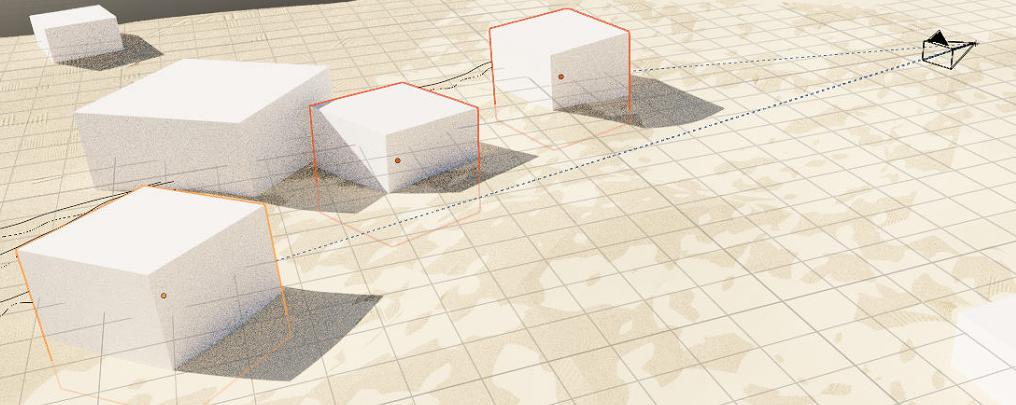}
    \caption{Image demo }
    \label{fig:demo}
\end{figure}
\noindent\textbf{Human in the Loop} Blender is designed for 3D editing, and it is intuitive and efficient to interact with 3D scenes through the GUI. Therefore, to allow users to control the objects in the scene precisely, we believe it is essential to preserve the user interface. With both visual and textual interactions available, users can select an object and then prompt BlenderGPT with a textual request. Figure~\ref{fig:demo} demonstrates an example of asking the camera to follow a snake's movement in the scene. Three snakes are shown as white cubic placeholders highlighted with orange sides in this coarse scene. The camera was originally following the snake far from the viewpoint. As there are multiple snakes in the scene, it could be challenging and inefficient to use only textual input to describe the desired object. Instead, the user can directly click on the snake object closest to the viewpoint with the textual input ``follow the selected object during the whole animation". 

\section{Experiment}
\label{sec:experiment}

We demonstrate the capabilities of \pipeline{} by rendering videos from generated Blender scenes. \\
\noindent\textbf{Implementation details} We choose Claude-3.5-Sonnet\,\cite{claude-3-5} as the model in Scene Codex and GPT-4o\,\cite{gpt4o} as the model in BlenderGPT. To avoid hallucinations and ensure relevance to the given prompts, we set the decoding temperature to 0. All methods are implemented using PyTorch 2.4.0 and executed on multiple NVIDIA H100 GPUs. 

\subsection{Qualitative Results}

\begin{figure*}[htbp]
    \centering
    \begin{subfigure}[h]{0.8\textwidth}
        \centering
        \includegraphics[width=\textwidth]{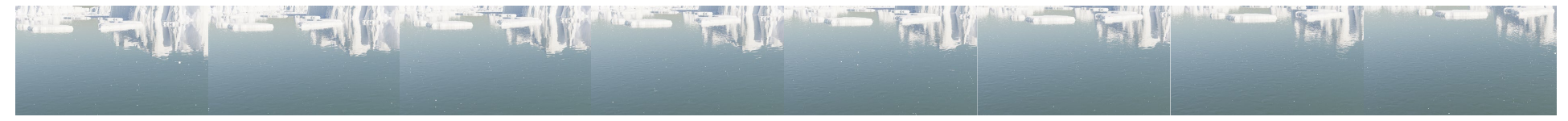}
        \captionsetup{justification=centering}
        \subcaption{Arctic: "The movement of sea ice"}
    \end{subfigure}

    \begin{subfigure}[h]{0.8\textwidth}
        \centering
        \includegraphics[width=\textwidth]{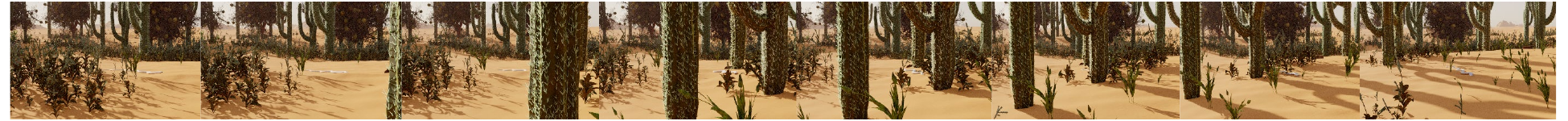}
        \captionsetup{justification=centering}
        \subcaption{Desert: "The snake in the desert"}
    \end{subfigure}

    \begin{subfigure}[h]{0.8\textwidth}
        \centering
        \includegraphics[width=\textwidth]{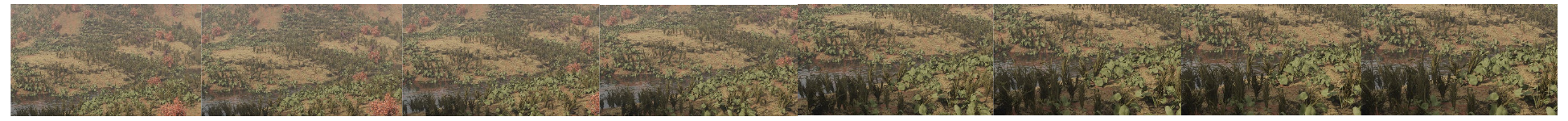}
        \captionsetup{justification=centering}
        \subcaption{Nature: "A Wide shot of a beautiful nature"}
    \end{subfigure}

    \begin{subfigure}[h]{0.8\textwidth}
        \centering
        \includegraphics[width=\textwidth]{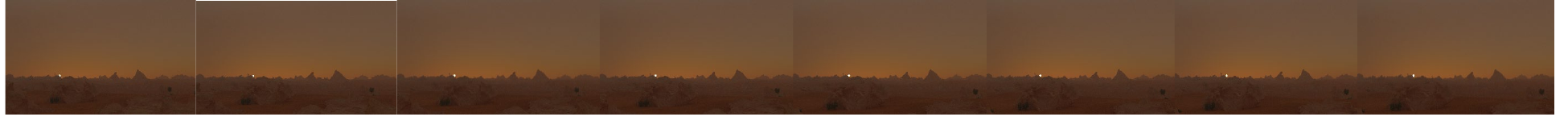}
        \captionsetup{justification=centering}
        \subcaption{Desert: "The sunset in the desert"}
    \end{subfigure}
    
    \begin{subfigure}[h]{0.8\textwidth}
        \centering
        \includegraphics[width=\textwidth]{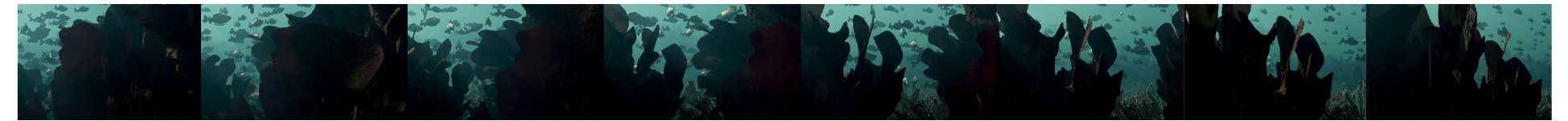}
        \captionsetup{justification=centering}
        \subcaption{Underwater: "The underwater sea world"}
    \end{subfigure}

    \caption{Five samples of eight key frames from generated videos with the corresponding textual input prompt.}
    \label{fig:result}
\end{figure*}

\begin{figure*}[htbp]
    \centering
    \begin{subfigure}[h]{0.8\textwidth}
        \centering
        \includegraphics[width=\textwidth]{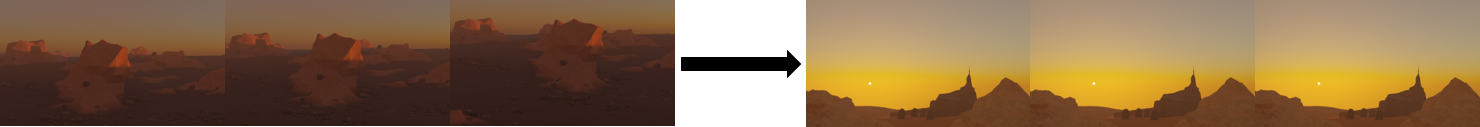}
        \captionsetup{justification=centering}
        \subcaption{"graveyard at sunset"}
        \label{fig:church_compare}
    \end{subfigure}

    \begin{subfigure}[h]{0.8\textwidth}
        \centering
        \includegraphics[width=\textwidth]{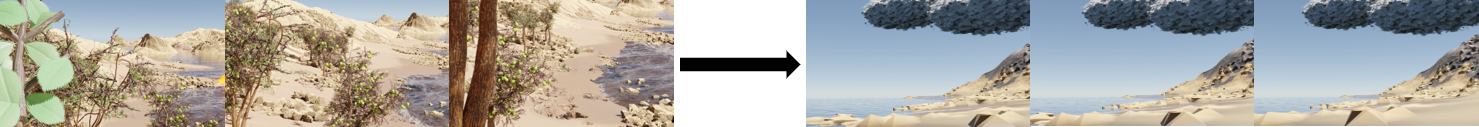}
        \captionsetup{justification=centering}
        \subcaption{"a relaxing scenery of beach view under cloudy sky"}
        \label{fig:beach_compare}
    \end{subfigure}

    \begin{subfigure}[h]{0.8\textwidth}
        \centering
        \includegraphics[width=\textwidth]{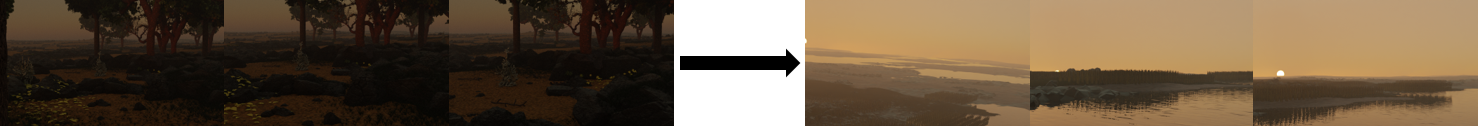}
        \captionsetup{justification=centering}
        \subcaption{"time lapse of a sunset sky in the countryside"}
        \label{fig:sunset_compare}
    \end{subfigure}
    
    \caption{\textbf{Left} Three key frames from rendered videos of the original scene output from Infinigen. \textbf{Right} Three key frames from rendered videos of the edited scene using \pipeline{}. Associated text prompts from VBench\,\cite{vbench} are used as the direct textual input.}
    \label{fig:compare}
\end{figure*}

In Figure~\ref{fig:result}, we present five samples of eight key frames from the rendered videos. We start with a base scene and provide additional prompts to update objects and details using \pipeline{}. After direct GUI-based revision such as moving camera positions or selecting desired subjects, the final output videos closely follow the textual prompts. 
In Figure~\ref{fig:compare}, we compare the rendered videos between the direct output videos from Infinigen and the results after editing with \pipeline{}. Note that we optimized Scene Codex specifically for two different tasks. Because of Infinigen's randomness, the direct output videos have a high probability of not focusing on the main subjects or may even fail to generate the requested assets. In contrast, with \pipeline{}, the user, acting as a ``director", can have more control over the scene and the output video. For example, as illustrated in Figure~\ref{fig:church_compare}, since Infinigen does not include a \emph{``graveyard"} in the asset, the camera is pointed in a random direction. However, using BlenderGPT, we generated a church and gravestones with fixed camera animation. 


\subsection{Quantitative Results}
\begin{table*}[ht]
\noindent
\small
\caption{\textbf{VBench result on Scenery category.} We compare our \pipeline{} with the other four T2V AI-based models in the Scenery category across eight dimensions. The best result in each dimension is highlighted in bold, and the second-best result is highlighted in Italics.}
\label{table:mainresult}
\centering
\scriptsize
\begin{tabular}{c|ccccccc}
\toprule
Models  & \makecell{Subject\\Consistency} & \makecell{Background\\Consistency} & \makecell{Motion\\Smoothness} & \makecell{Dynamic\\Degree} & \makecell{Aesthetic\\Quality} & \makecell{Imaging\\Quality}  \\
\midrule
LaVie & \textbf{97.27} & \textbf{97.06} & \textit{97.58} & 6.40 & \textbf{51.76} & \textbf{63.86} \\
ModelScope & \textit{94.88} & \textit{95.57} & 97.03 & 26.00 & 48.57 & 57.49  \\
VideoCrafter1 & 89.67 & 92.86 & 94.17 & \textit{51.80} & 43.06 & 58.98 \\
CogVideo & 95.27 & 95.46 & \textit{97.58} & 13.20 & 46.72 & 40.49  \\
\midrule
Scene Copilot (Ours) & 93.10 & 95.17 & \textbf{99.36} & \textbf{66.67} & \textit{49.24} & \textit{62.10} \\

\bottomrule
\end{tabular}
\vspace*{-0.2cm}

\end{table*}
To quantitatively evaluate the output videos, we applied VBench\,\cite{vbench}, a comprehensive video evaluation benchmark with multiple dimensions, and targeted on 100 textual prompts in \emph{scenery} category as Infinigen's major strength lies in creating natural photorealistic scenes. We created scenes using VBench text prompts as the only input and edited details with BlenderGPT. The average editing time was approximately 20 minutes. Table~\ref{table:mainresult} summarizes our main results compared with other AI-driven (transformer/diffusion-based) video generation models including LaVie\,\cite{wang2023lavie}, ModelScope \,\cite{wang2023modelscope} CogVideo\,\cite{hong2022cogvideo}, and VideoCrafter1\,\cite{chen2023videocrafter1}. Overall, our generated videos show competitive results with similar scores compared to other autoregressive text-to-video(T2V) models. Specifically, \pipeline{} achieves the best results in Motion Smoothness and Dynamic Degree and the second-best in Aesthetic Quality and Imaging Quality. Both Motion Smoothness and Dynamic Degree evaluate whether the motion in the generated video is smooth and follows the physical law of the real world \,\cite{vbench}. Achieving the best results in these two dimensions and second-best in frame-wise quality demonstrates great potential and feasibility in generating video through Infinigen, enhanced by our \pipeline{}.



\subsection{Ablation}
\begin{figure}[h]
    \centering
    \begin{minipage}[t]{0.21\textwidth}
        \centering
        \caption{The evaluation result of prompting RAG database Infinigen and few-shot examples. Based on the result, Using both of them are valuable and vital for our Scene Codex.}
        \label{table:ablation}
        \setlength{\tabcolsep}{2pt}
        \footnotesize 
        \begin{tabular}{c|cc|c}
            \toprule
            & \makecell{RAG} & \makecell{Few-Shot} & \makecell{ER@1} \\
            \midrule
            1 &  &  & 0.0$\%$ \\
            2 &  & \checkmark  & 2.0$\%$ \\
            3 & \checkmark &  & 20.0$\%$ \\
            4 & \checkmark & \checkmark  & \textbf{43.2$\%$} \\
            \midrule
        \end{tabular}

        \vspace*{-0.2cm}
    \end{minipage}%
    \hfill
    \begin{minipage}[t]{0.265\textwidth}
        \centering
        \vspace{0pt} 
        \begin{tikzpicture}
            \begin{axis}[
                xlabel={Video length (min)},
                ylabel={Success Rate (\%)},
                ymin=0, ymax=110,
                xlabel near ticks,
                ylabel near ticks,
                xlabel shift={-3pt},
                ylabel shift={-5pt},
                symbolic x coords={1, 2, 5, 10, 30},
                xtick=data,
                ytick={0, 20, 40, 60, 80, 100},
                width=\linewidth,
                height=\textwidth,
                enlargelimits=0.05,
                grid=both,
                mark options={solid},
                xlabel style={font=\footnotesize }, 
                ylabel style={font=\footnotesize },
                line width=1pt,
                legend style={font=\footnotesize, at={(1,0.7)}, inner sep=0pt, column sep=1pt},
                xticklabel style={font=\footnotesize, rotate=0}
            ]
                \addplot[mark=*, color=blue] coordinates {
                    (1, 74.1)
                    (2, 50)
                    (5, 9.2)
                    (10, 5.3)
                    (30, 0)
                };
                \addlegendentry{Infinigen}
                
                \addplot[mark=o, color=red, dotted] coordinates {
                    (1, 100)
                    (2, 100)
                    (5, 100)
                    (10, 100)
                    (30, 100)
                };
                \addlegendentry{Ours}
                
            \end{axis}
        \end{tikzpicture}
        \caption{Video generation success rate comparison between Infinigen and Ours across different video lengths. Infinigen's performance decreases as video length increases.}
        \label{fig:longvideo}
    \end{minipage}
\end{figure}
As elaborated in Section \ref{sec:codex}, our Scene Codex combines RAG and few-shot examples to generate the Infinigen commands. We investigate the effect of each component on the generated commands. We used textual prompts from VBench as input and measured the Executable Rate (ER@1) by calculating the proportion of generated commands that are executable, to evaluate the usability of our Scene Codex \,\cite{zhou2024scenex, codex}. Table~\ref{table:ablation} shows the results ranging from using only the LLM to the final model with both techniques applied. It is evident that both the RAG database of Infinigen and few-shot examples are valuable to our Scene Codex. Without including any references, it is impossible to directly generate any executable Infinigen commands. With few-shot examples, Scene Codex imitated examples and produced 2.0$\%$ runnable commands. Extra context from RAG database helps the LLMs understand the code relationships, boosting the runnable commands to 20.0$\%$. Finally, when we include both components in our Scene Codex, the ER@1 reaches 43.2$\%$. This result demonstrates the value of including additional information when prompting LLMs.

\subsection{Long-Video Generation}

Many models have demonstrated the ability to generate high-quality short-duration videos. However, generating videos spanning minute-level or even hour-level durations remains a challenge, particularly with respect to maintaining object integrity and adhering to physical laws \cite{blattmann2023align,blattmann2023stable,ho2022video}. One advantage of generating videos from a pre-constructed scene is the consistency of the objects within that scene. This consistency allows for the creation of videos of arbitrary length without concern for the preservation of object integrity or the enforcement of the physical laws governing the generated world. We evaluated the long-video generation capability of Infinigen and Infinigen with \pipeline{}. We generated 12 scenes in each Infinigen-provided procedural scene setting and calculated the success rate at  different video lengths, as shown in Figure~\ref{fig:longvideo}. From the figure, Infinigen is capable of generating one-minute-long videos with a $74.1\%$ success rate, but the success rate decreases as the requested video length increases. This is because Infinigen has to find an existing camera path spanning the whole video length before it can create the scene. In contrast, benefiting from generating scenes with human-in-the-loop, users can start with a one-second-long scene and customize the camera animation for videos of arbitrary length.

\section{Conclusion}
\label{sec:conclusion}
In this work, we present \pipeline{}, an automatic pipeline combining procedural scene generation and human-centered editing with additional procedural assets. \pipeline{} consists of Scene Codex and BlenderGPT. Utilizing remarkable LLMs, Scene Codex converts user prompts into an Infinigen-based Python command to create the base 3D scene. BlenderGPT then helps users accurately revise and update the 3D scene with visual and textual human input. Additionally, our procedural dataset provides more than 300 extra geometric and material assets to make \pipeline{} more versatile.
Supported by unprecedented LLMs and our procedural dataset, Scene Codex, and BlenderGPT could help users control the 3D scene generation with high precision.  
We hope \pipeline{} can lower the bound of creating high-quality, dynamic, and controllable 3D scenes.

\noindent\textbf{Limitations} Although \pipeline{} demonstrated promising ability in supporting users' 3D scene editing, we would like to acknowledge several assumptions and limitations: 1) Because we introduced LLMs in \pipeline{}, the hallucination of LLMs is inevitable, resulting in advanced programming knowledge from user to help correct issues and mistakes. This issue may shield the power of introducing LLMs in the framework, but it could be alleviated with advancements of more powerful LLMs. 2) The amount of procedural objects in our datasets is limited, compared with other 3D model datasets in other formats such as mesh and point cloud. Creating procedural geometry objects is time-consuming and most of them are proprietary. We plan to continuously include new assets in our datasets, and we believe it is possible to generate procedural geometry assets with multi-modal LLMs and neural networks.

\clearpage 
{
    \small
    \bibliographystyle{ieeenat_fullname}
    \bibliography{7_reference}

\begin{thebibliography}{75}
\providecommand{\natexlab}[1]{#1}
\providecommand{\url}[1]{\texttt{#1}}
\expandafter\ifx\csname urlstyle\endcsname\relax
  \providecommand{\doi}[1]{doi: #1}\else
  \providecommand{\doi}{doi: \begingroup \urlstyle{rm}\Url}\fi

\bibitem[Anthropic(2024)]{claude-3-5}
Anthropic.
\newblock Claude 3.5 sonnet, 2024.

\bibitem[Blattmann et~al.(2023{\natexlab{a}})Blattmann, Dockhorn, Kulal, Mendelevitch, Kilian, Lorenz, Levi, English, Voleti, Letts, et~al.]{blattmann2023stable}
Andreas Blattmann, Tim Dockhorn, Sumith Kulal, Daniel Mendelevitch, Maciej Kilian, Dominik Lorenz, Yam Levi, Zion English, Vikram Voleti, Adam Letts, et~al.
\newblock Stable video diffusion: Scaling latent video diffusion models to large datasets.
\newblock \emph{arXiv preprint arXiv:2311.15127}, 2023{\natexlab{a}}.

\bibitem[Blattmann et~al.(2023{\natexlab{b}})Blattmann, Rombach, Ling, Dockhorn, Kim, Fidler, and Kreis]{blattmann2023align}
Andreas Blattmann, Robin Rombach, Huan Ling, Tim Dockhorn, Seung~Wook Kim, Sanja Fidler, and Karsten Kreis.
\newblock Align your latents: High-resolution video synthesis with latent diffusion models.
\newblock In \emph{Proceedings of the IEEE/CVF Conference on Computer Vision and Pattern Recognition}, pages 22563--22575, 2023{\natexlab{b}}.

\bibitem[BrendanParmer(2023)]{node2python}
carls3d BrendanParmer.
\newblock Nodetopython.
\newblock \url{https://github.com/BrendanParmer/NodeToPython}, 2023.

\bibitem[Brooks et~al.(2024)Brooks, Peebles, Homes, DePue, Guo, Jing, Schnurr, Taylor, Luhman, Luhman, Ng, Wang, and Ramesh]{sora}
Tim Brooks, Bill Peebles, Connor Homes, Will DePue, Yufei Guo, Li Jing, David Schnurr, Joe Taylor, Troy Luhman, Eric Luhman, Clarence Wing~Yin Ng, Ricky Wang, and Aditya Ramesh.
\newblock Video generation models as world simulators.
\newblock 2024.

\bibitem[Brown(2020)]{brown2020language}
Tom~B Brown.
\newblock Language models are few-shot learners.
\newblock \emph{arXiv preprint arXiv:2005.14165}, 2020.

\bibitem[Butler et~al.(2012)Butler, Wulff, Stanley, and Black]{butler2012naturalistic}
Daniel~J Butler, Jonas Wulff, Garrett~B Stanley, and Michael~J Black.
\newblock A naturalistic open source movie for optical flow evaluation.
\newblock In \emph{Computer Vision--ECCV 2012: 12th European Conference on Computer Vision, Florence, Italy, October 7-13, 2012, Proceedings, Part VI 12}, pages 611--625. Springer, 2012.

\bibitem[Caesar et~al.(2020)Caesar, Bankiti, Lang, Vora, Liong, Xu, Krishnan, Pan, Baldan, and Beijbom]{caesar2020nuscenesmultimodaldatasetautonomous}
Holger Caesar, Varun Bankiti, Alex~H. Lang, Sourabh Vora, Venice~Erin Liong, Qiang Xu, Anush Krishnan, Yu Pan, Giancarlo Baldan, and Oscar Beijbom.
\newblock nuscenes: A multimodal dataset for autonomous driving, 2020.

\bibitem[Cao et~al.(2024)Cao, Yihan, Xu, and Xu]{cao2024coda}
Yang Cao, Zeng Yihan, Hang Xu, and Dan Xu.
\newblock Coda: Collaborative novel box discovery and cross-modal alignment for open-vocabulary 3d object detection.
\newblock \emph{Advances in Neural Information Processing Systems}, 36, 2024.

\bibitem[Chen et~al.(2023)Chen, Xia, He, Zhang, Cun, Yang, Xing, Liu, Chen, Wang, Weng, and Shan]{chen2023videocrafter1}
Haoxin Chen, Menghan Xia, Yingqing He, Yong Zhang, Xiaodong Cun, Shaoshu Yang, Jinbo Xing, Yaofang Liu, Qifeng Chen, Xintao Wang, Chao Weng, and Ying Shan.
\newblock Videocrafter1: Open diffusion models for high-quality video generation.
\newblock \emph{arXiv preprint arXiv:2310.19512}, 2023.

\bibitem[Chen et~al.(2021)Chen, Tworek, Jun, Yuan, Pinto, Kaplan, Edwards, Burda, Joseph, Brockman, et~al.]{codex}
Mark Chen, Jerry Tworek, Heewoo Jun, Qiming Yuan, Henrique Ponde De~Oliveira Pinto, Jared Kaplan, Harri Edwards, Yuri Burda, Nicholas Joseph, Greg Brockman, et~al.
\newblock Evaluating large language models trained on code.
\newblock \emph{arXiv preprint arXiv:2107.03374}, 2021.

\bibitem[Community(2018)]{blender}
Blender~Online Community.
\newblock \emph{Blender - a 3D modelling and rendering package}.
\newblock Blender Foundation, Stichting Blender Foundation, Amsterdam, 2018.

\bibitem[Deitke et~al.(2022)Deitke, VanderBilt, Herrasti, Weihs, Ehsani, Salvador, Han, Kolve, Kembhavi, and Mottaghi]{deitke2022️}
Matt Deitke, Eli VanderBilt, Alvaro Herrasti, Luca Weihs, Kiana Ehsani, Jordi Salvador, Winson Han, Eric Kolve, Aniruddha Kembhavi, and Roozbeh Mottaghi.
\newblock Procthor: Large-scale embodied ai using procedural generation.
\newblock \emph{Advances in Neural Information Processing Systems}, 35:\penalty0 5982--5994, 2022.

\bibitem[Ding et~al.(2023)Ding, Yang, Xue, Zhang, Bai, and Qi]{ding2023pla}
Runyu Ding, Jihan Yang, Chuhui Xue, Wenqing Zhang, Song Bai, and Xiaojuan Qi.
\newblock Pla: Language-driven open-vocabulary 3d scene understanding.
\newblock In \emph{Proceedings of the IEEE/CVF conference on computer vision and pattern recognition}, pages 7010--7019, 2023.

\bibitem[Douze et~al.(2024)Douze, Guzhva, Deng, Johnson, Szilvasy, Mazaré, Lomeli, Hosseini, and Jégou]{faiss}
Matthijs Douze, Alexandr Guzhva, Chengqi Deng, Jeff Johnson, Gergely Szilvasy, Pierre-Emmanuel Mazaré, Maria Lomeli, Lucas Hosseini, and Hervé Jégou.
\newblock The faiss library, 2024.

\bibitem[Fabbri et~al.(2021)Fabbri, Bras{\'o}, Maugeri, Cetintas, Gasparini, O{\v{s}}ep, Calderara, Leal-Taix{\'e}, and Cucchiara]{fabbri2021motsynth}
Matteo Fabbri, Guillem Bras{\'o}, Gianluca Maugeri, Orcun Cetintas, Riccardo Gasparini, Aljo{\v{s}}a O{\v{s}}ep, Simone Calderara, Laura Leal-Taix{\'e}, and Rita Cucchiara.
\newblock Motsynth: How can synthetic data help pedestrian detection and tracking?
\newblock In \emph{Proceedings of the IEEE/CVF International Conference on Computer Vision}, pages 10849--10859, 2021.

\bibitem[Fang et~al.(2024)Fang, Miao, Srivastav, Liu, Zhang, Fang, Asmita, Tsang, Nazari, Wang, and Homayoun]{llmcode1}
Chongzhou Fang, Ning Miao, Shaurya Srivastav, Jialin Liu, Ruoyu Zhang, Ruijie Fang, Asmita, Ryan Tsang, Najmeh Nazari, Han Wang, and Houman Homayoun.
\newblock Large language models for code analysis: Do {LLMs} really do their job?
\newblock In \emph{33rd USENIX Security Symposium (USENIX Security 24)}, pages 829--846, Philadelphia, PA, 2024. USENIX Association.

\bibitem[Fang et~al.(2025)Fang, Wang, Tsai, Yang, Ding, Zhou, and Yang]{fang2025chat}
Shuangkang Fang, Yufeng Wang, Yi-Hsuan Tsai, Yi Yang, Wenrui Ding, Shuchang Zhou, and Ming-Hsuan Yang.
\newblock Chat-edit-3d: Interactive 3d scene editing via text prompts.
\newblock In \emph{European Conference on Computer Vision}, pages 199--216. Springer, 2025.

\bibitem[Fridman et~al.(2024)Fridman, Abecasis, Kasten, and Dekel]{fridman2024scenescape}
Rafail Fridman, Amit Abecasis, Yoni Kasten, and Tali Dekel.
\newblock Scenescape: Text-driven consistent scene generation.
\newblock \emph{Advances in Neural Information Processing Systems}, 36, 2024.

\bibitem[Fu et~al.(2024)Fu, Liu, Chen, Nie, and Xiong]{fu2024scene}
Rao Fu, Jingyu Liu, Xilun Chen, Yixin Nie, and Wenhan Xiong.
\newblock Scene-llm: Extending language model for 3d visual understanding and reasoning.
\newblock \emph{arXiv preprint arXiv:2403.11401}, 2024.

\bibitem[Gao et~al.(2024)Gao, Xiong, Gao, Jia, Pan, Bi, Dai, Sun, Wang, and Wang]{rag_survey}
Yunfan Gao, Yun Xiong, Xinyu Gao, Kangxiang Jia, Jinliu Pan, Yuxi Bi, Yi Dai, Jiawei Sun, Meng Wang, and Haofen Wang.
\newblock Retrieval-augmented generation for large language models: A survey, 2024.

\bibitem[Gasch et~al.(2022)Gasch, Sotoca, Chover, Remolar, and Rebollo]{gasch2022procedural}
Cristina Gasch, Jos{\'e}~Mart{\'\i}nez Sotoca, Miguel Chover, Inmaculada Remolar, and Cristina Rebollo.
\newblock Procedural modeling of plant ecosystems maximizing vegetation cover.
\newblock \emph{Multimedia Tools and Applications}, 81\penalty0 (12):\penalty0 16195--16217, 2022.

\bibitem[Henschel et~al.(2024)Henschel, Khachatryan, Hayrapetyan, Poghosyan, Tadevosyan, Wang, Navasardyan, and Shi]{streamingt2v}
Roberto Henschel, Levon Khachatryan, Daniil Hayrapetyan, Hayk Poghosyan, Vahram Tadevosyan, Zhangyang Wang, Shant Navasardyan, and Humphrey Shi.
\newblock Streamingt2v: Consistent, dynamic, and extendable long video generation from text.
\newblock \emph{arXiv preprint arXiv:2403.14773}, 2024.

\bibitem[Ho et~al.(2022)Ho, Salimans, Gritsenko, Chan, Norouzi, and Fleet]{ho2022video}
Jonathan Ho, Tim Salimans, Alexey Gritsenko, William Chan, Mohammad Norouzi, and David~J Fleet.
\newblock Video diffusion models.
\newblock \emph{Advances in Neural Information Processing Systems}, 35:\penalty0 8633--8646, 2022.

\bibitem[Hong et~al.(2022)Hong, Ding, Zheng, Liu, and Tang]{hong2022cogvideo}
Wenyi Hong, Ming Ding, Wendi Zheng, Xinghan Liu, and Jie Tang.
\newblock {CogVideo}: Large-scale pretraining for text-to-video generation via transformers.
\newblock \emph{arXiv preprint arXiv:2205.15868}, 2022.

\bibitem[Hu et~al.(2022)Hu, He, Deschaintre, Dorsey, and Rushmeier]{hu2022inverse}
Yiwei Hu, Chengan He, Valentin Deschaintre, Julie Dorsey, and Holly Rushmeier.
\newblock An inverse procedural modeling pipeline for svbrdf maps.
\newblock \emph{ACM Transactions on Graphics (TOG)}, 41\penalty0 (2):\penalty0 1--17, 2022.

\bibitem[Hu et~al.(2023)Hu, Guerrero, Hasan, Rushmeier, and Deschaintre]{hu2023generating}
Yiwei Hu, Paul Guerrero, Milos Hasan, Holly Rushmeier, and Valentin Deschaintre.
\newblock Generating procedural materials from text or image prompts.
\newblock In \emph{ACM SIGGRAPH 2023 Conference Proceedings}, pages 1--11, 2023.

\bibitem[Hu et~al.(2024)Hu, Iscen, Jain, Kipf, Yue, Ross, Schmid, and Fathi]{hu2024scenecraft}
Ziniu Hu, Ahmet Iscen, Aashi Jain, Thomas Kipf, Yisong Yue, David~A Ross, Cordelia Schmid, and Alireza Fathi.
\newblock Scenecraft: An llm agent for synthesizing 3d scenes as blender code.
\newblock In \emph{Forty-first International Conference on Machine Learning}, 2024.

\bibitem[Huang et~al.(2024)Huang, He, Yu, Zhang, Si, Jiang, Zhang, Wu, Jin, Chanpaisit, Wang, Chen, Wang, Lin, Qiao, and Liu]{vbench}
Ziqi Huang, Yinan He, Jiashuo Yu, Fan Zhang, Chenyang Si, Yuming Jiang, Yuanhan Zhang, Tianxing Wu, Qingyang Jin, Nattapol Chanpaisit, Yaohui Wang, Xinyuan Chen, Limin Wang, Dahua Lin, Yu Qiao, and Ziwei Liu.
\newblock {VBench}: Comprehensive benchmark suite for video generative models.
\newblock In \emph{Proceedings of the IEEE/CVF Conference on Computer Vision and Pattern Recognition}, 2024.

\bibitem[Jain et~al.(2021)Jain, Tancik, and Abbeel]{jain2021putting}
Ajay Jain, Matthew Tancik, and Pieter Abbeel.
\newblock Putting nerf on a diet: Semantically consistent few-shot view synthesis.
\newblock In \emph{Proceedings of the IEEE/CVF International Conference on Computer Vision}, pages 5885--5894, 2021.

\bibitem[Jain et~al.(2022)Jain, Mildenhall, Barron, Abbeel, and Poole]{jain2022zero}
Ajay Jain, Ben Mildenhall, Jonathan~T Barron, Pieter Abbeel, and Ben Poole.
\newblock Zero-shot text-guided object generation with dream fields.
\newblock In \emph{Proceedings of the IEEE/CVF conference on computer vision and pattern recognition}, pages 867--876, 2022.

\bibitem[Jun and Nichol(2023)]{jun2023shape}
Heewoo Jun and Alex Nichol.
\newblock Shap-e: Generating conditional 3d implicit functions, 2023.

\bibitem[Khalid et~al.(2023)Khalid, Xie, Belilovsky, and Popa]{khalid2023clip}
N Khalid, T Xie, E Belilovsky, and T Popa.
\newblock \emph{Clip-mesh: Generating textured meshes from text using pretrained image-text models}.
\newblock PhD thesis, Concordia University Montr{\'e}al, Qu{\'e}bec, Canada, 2023.

\bibitem[Kuaishou(2024)]{klingai}
Kuaishou.
\newblock Kling {AI}.
\newblock \url{https://klingai.kuaishou.com}, 2024.

\bibitem[Lee and Chang(2022)]{lee2022understanding}
Han-Hung Lee and Angel~X Chang.
\newblock Understanding pure clip guidance for voxel grid nerf models.
\newblock \emph{arXiv preprint arXiv:2209.15172}, 2022.

\bibitem[Lewis et~al.(2021)Lewis, Perez, Piktus, Petroni, Karpukhin, Goyal, Küttler, Lewis, tau Yih, Rocktäschel, Riedel, and Kiela]{rag}
Patrick Lewis, Ethan Perez, Aleksandra Piktus, Fabio Petroni, Vladimir Karpukhin, Naman Goyal, Heinrich Küttler, Mike Lewis, Wen tau Yih, Tim Rocktäschel, Sebastian Riedel, and Douwe Kiela.
\newblock Retrieval-augmented generation for knowledge-intensive nlp tasks, 2021.

\bibitem[Li et~al.(2024)Li, Wang, Zheng, and Zhang]{li2024looglelongcontextlanguagemodels}
Jiaqi Li, Mengmeng Wang, Zilong Zheng, and Muhan Zhang.
\newblock Loogle: Can long-context language models understand long contexts?, 2024.

\bibitem[Liu et~al.(2024)Liu, Huang, Hou, Wang, Yin, Gong, Gao, and Ouyang]{liu2024uni3d}
Dingning Liu, Xiaoshui Huang, Yuenan Hou, Zhihui Wang, Zhenfei Yin, Yongshun Gong, Peng Gao, and Wanli Ouyang.
\newblock Uni3d-llm: Unifying point cloud perception, generation and editing with large language models.
\newblock \emph{arXiv preprint arXiv:2402.03327}, 2024.

\bibitem[Liu et~al.(2021{\natexlab{a}})Liu, Shen, Zhang, Dolan, Carin, and Chen]{liu2021makes}
Jiachang Liu, Dinghan Shen, Yizhe Zhang, Bill Dolan, Lawrence Carin, and Weizhu Chen.
\newblock What makes good in-context examples for gpt-$3 $?
\newblock \emph{arXiv preprint arXiv:2101.06804}, 2021{\natexlab{a}}.

\bibitem[Liu et~al.(2021{\natexlab{b}})Liu, Snodgrass, Khalifa, Risi, Yannakakis, and Togelius]{liu2021deep}
Jialin Liu, Sam Snodgrass, Ahmed Khalifa, Sebastian Risi, Georgios~N Yannakakis, and Julian Togelius.
\newblock \emph{Deep learning for procedural content generation}.
\newblock Springer, 2021{\natexlab{b}}.

\bibitem[Lu et~al.(2024)Lu, Chen, Williamson, Chen, Zhao, Chow, Ikemura, Kim, Pouli, Patel, et~al.]{lu2024multimodal}
Ming~Y Lu, Bowen Chen, Drew~FK Williamson, Richard~J Chen, Melissa Zhao, Aaron~K Chow, Kenji Ikemura, Ahrong Kim, Dimitra Pouli, Ankush Patel, et~al.
\newblock A multimodal generative ai copilot for human pathology.
\newblock \emph{Nature}, pages 1--3, 2024.

\bibitem[Lu et~al.(2023)Lu, Chang, Jing, Boularias, and Bekris]{lu2023ovir}
Shiyang Lu, Haonan Chang, Eric~Pu Jing, Abdeslam Boularias, and Kostas Bekris.
\newblock Ovir-3d: Open-vocabulary 3d instance retrieval without training on 3d data.
\newblock In \emph{Conference on Robot Learning}, pages 1610--1620. PMLR, 2023.

\bibitem[Ma et~al.(2024)Ma, Bhalgat, Smart, Chen, Li, Ding, Gu, Chen, Peng, Bian, et~al.]{ma2024llms}
Xianzheng Ma, Yash Bhalgat, Brandon Smart, Shuai Chen, Xinghui Li, Jian Ding, Jindong Gu, Dave~Zhenyu Chen, Songyou Peng, Jia-Wang Bian, et~al.
\newblock When llms step into the 3d world: A survey and meta-analysis of 3d tasks via multi-modal large language models.
\newblock \emph{arXiv preprint arXiv:2405.10255}, 2024.

\bibitem[Mao et~al.(2021)Mao, Niu, Jiang, Liang, Chen, Liang, Li, Ye, Zhang, Li, et~al.]{mao2021one}
Jiageng Mao, Minzhe Niu, Chenhan Jiang, Hanxue Liang, Jingheng Chen, Xiaodan Liang, Yamin Li, Chaoqiang Ye, Wei Zhang, Zhenguo Li, et~al.
\newblock One million scenes for autonomous driving: Once dataset.
\newblock \emph{arXiv preprint arXiv:2106.11037}, 2021.

\bibitem[Min et~al.(2022)Min, Lyu, Holtzman, Artetxe, Lewis, Hajishirzi, and Zettlemoyer]{min2022rethinking}
Sewon Min, Xinxi Lyu, Ari Holtzman, Mikel Artetxe, Mike Lewis, Hannaneh Hajishirzi, and Luke Zettlemoyer.
\newblock Rethinking the role of demonstrations: What makes in-context learning work?
\newblock \emph{arXiv preprint arXiv:2202.12837}, 2022.

\bibitem[OpenAI(2024{\natexlab{a}})]{gpt4o}
OpenAI.
\newblock Hello gpt-4o.
\newblock \emph{OpenAI Blog}, 2024{\natexlab{a}}.

\bibitem[OpenAI(2024{\natexlab{b}})]{gpt4omini}
OpenAI.
\newblock Gpt-4o mini: advancing cost-efficient intelligence.
\newblock \emph{OpenAI Blog}, 2024{\natexlab{b}}.

\bibitem[Raistrick et~al.(2023)Raistrick, Lipson, Ma, Mei, Wang, Zuo, Kayan, Wen, Han, Wang, Newell, Law, Goyal, Yang, and Deng]{infinigen2023infinite}
Alexander Raistrick, Lahav Lipson, Zeyu Ma, Lingjie Mei, Mingzhe Wang, Yiming Zuo, Karhan Kayan, Hongyu Wen, Beining Han, Yihan Wang, Alejandro Newell, Hei Law, Ankit Goyal, Kaiyu Yang, and Jia Deng.
\newblock Infinite photorealistic worlds using procedural generation.
\newblock In \emph{Proceedings of the IEEE/CVF Conference on Computer Vision and Pattern Recognition}, pages 12630--12641, 2023.

\bibitem[Raistrick et~al.(2024)Raistrick, Mei, Kayan, Yan, Zuo, Han, Wen, Parakh, Alexandropoulos, Lipson, Ma, and Deng]{infinigen2024indoors}
Alexander Raistrick, Lingjie Mei, Karhan Kayan, David Yan, Yiming Zuo, Beining Han, Hongyu Wen, Meenal Parakh, Stamatis Alexandropoulos, Lahav Lipson, Zeyu Ma, and Jia Deng.
\newblock Infinigen indoors: Photorealistic indoor scenes using procedural generation.
\newblock In \emph{Proceedings of the IEEE/CVF Conference on Computer Vision and Pattern Recognition (CVPR)}, pages 21783--21794, 2024.

\bibitem[Risi and Togelius(2020)]{risi2020increasing}
Sebastian Risi and Julian Togelius.
\newblock Increasing generality in machine learning through procedural content generation.
\newblock \emph{Nature Machine Intelligence}, 2\penalty0 (8):\penalty0 428--436, 2020.

\bibitem[Roberts et~al.(2021)Roberts, Ramapuram, Ranjan, Kumar, Bautista, Paczan, Webb, and Susskind]{roberts2021hypersim}
Mike Roberts, Jason Ramapuram, Anurag Ranjan, Atulit Kumar, Miguel~Angel Bautista, Nathan Paczan, Russ Webb, and Joshua~M Susskind.
\newblock Hypersim: A photorealistic synthetic dataset for holistic indoor scene understanding.
\newblock In \emph{Proceedings of the IEEE/CVF international conference on computer vision}, pages 10912--10922, 2021.

\bibitem[Rozière et~al.(2024)Rozière, Gehring, Gloeckle, Sootla, Gat, Tan, Adi, Liu, Sauvestre, Remez, Rapin, Kozhevnikov, Evtimov, Bitton, Bhatt, Ferrer, Grattafiori, Xiong, Défossez, Copet, Azhar, Touvron, Martin, Usunier, Scialom, and Synnaeve]{codellama}
Baptiste Rozière, Jonas Gehring, Fabian Gloeckle, Sten Sootla, Itai Gat, Xiaoqing~Ellen Tan, Yossi Adi, Jingyu Liu, Romain Sauvestre, Tal Remez, Jérémy Rapin, Artyom Kozhevnikov, Ivan Evtimov, Joanna Bitton, Manish Bhatt, Cristian~Canton Ferrer, Aaron Grattafiori, Wenhan Xiong, Alexandre Défossez, Jade Copet, Faisal Azhar, Hugo Touvron, Louis Martin, Nicolas Usunier, Thomas Scialom, and Gabriel Synnaeve.
\newblock Code llama: Open foundation models for code, 2024.

\bibitem[Shaker et~al.(2016)Shaker, Togelius, and Nelson]{shaker2016procedural}
Noor Shaker, Julian Togelius, and Mark~J Nelson.
\newblock \emph{Procedural content generation in games}.
\newblock Springer, 2016.

\bibitem[Shi et~al.(2020)Shi, Li, Ha{\v{s}}an, Sunkavalli, Boubekeur, Mech, and Matusik]{shi2020match}
Liang Shi, Beichen Li, Milo{\v{s}} Ha{\v{s}}an, Kalyan Sunkavalli, Tamy Boubekeur, Radomir Mech, and Wojciech Matusik.
\newblock Match: Differentiable material graphs for procedural material capture.
\newblock \emph{ACM Transactions on Graphics (TOG)}, 39\penalty0 (6):\penalty0 1--15, 2020.

\bibitem[Sun et~al.(2024{\natexlab{a}})Sun, Han, Deng, Wang, Qin, and Gould]{sun20243dgptprocedural3dmodeling}
Chunyi Sun, Junlin Han, Weijian Deng, Xinlong Wang, Zishan Qin, and Stephen Gould.
\newblock 3d-gpt: Procedural 3d modeling with large language models, 2024{\natexlab{a}}.

\bibitem[Sun et~al.(2024{\natexlab{b}})Sun, Zhang, Shah, Sun, Zhang, Li, Duan, Wei, and Ranjan]{sun2024sora}
Rui Sun, Yumin Zhang, Tejal Shah, Jiahao Sun, Shuoying Zhang, Wenqi Li, Haoran Duan, Bo Wei, and Rajiv Ranjan.
\newblock From sora what we can see: A survey of text-to-video generation.
\newblock \emph{arXiv preprint arXiv:2405.10674}, 2024{\natexlab{b}}.

\bibitem[Tang et~al.(2024)Tang, Han, Li, Yu, Hao, Hu, and Chen]{tang2024minigpt}
Yuan Tang, Xu Han, Xianzhi Li, Qiao Yu, Yixue Hao, Long Hu, and Min Chen.
\newblock Minigpt-3d: Efficiently aligning 3d point clouds with large language models using 2d priors.
\newblock In \emph{Proceedings of the 32nd ACM International Conference on Multimedia}, pages 6617--6626, 2024.

\bibitem[Vidanapathirana et~al.(2021)Vidanapathirana, Wu, Furukawa, Chang, and Savva]{vidanapathirana2021plan2scene}
Madhawa Vidanapathirana, Qirui Wu, Yasutaka Furukawa, Angel~X Chang, and Manolis Savva.
\newblock Plan2scene: Converting floorplans to 3d scenes.
\newblock In \emph{Proceedings of the IEEE/CVF Conference on Computer Vision and Pattern Recognition}, pages 10733--10742, 2021.

\bibitem[Wang et~al.(2023{\natexlab{a}})Wang, Yuan, Chen, Zhang, Wang, and Zhang]{wang2023modelscope}
Jiuniu Wang, Hangjie Yuan, Dayou Chen, Yingya Zhang, Xiang Wang, and Shiwei Zhang.
\newblock Modelscope text-to-video technical report.
\newblock \emph{arXiv preprint arXiv:2308.06571}, 2023{\natexlab{a}}.

\bibitem[Wang et~al.(2024)Wang, Mao, Zhu, Xu, Lyu, Li, Chen, Zhang, Chen, Xue, Liu, Lu, Lin, and Pang]{embodiedscan}
Tai Wang, Xiaohan Mao, Chenming Zhu, Runsen Xu, Ruiyuan Lyu, Peisen Li, Xiao Chen, Wenwei Zhang, Kai Chen, Tianfan Xue, Xihui Liu, Cewu Lu, Dahua Lin, and Jiangmiao Pang.
\newblock Embodiedscan: A holistic multi-modal 3d perception suite towards embodied ai.
\newblock In \emph{IEEE Conference on Computer Vision and Pattern Recognition (CVPR)}, 2024.

\bibitem[Wang et~al.(2023{\natexlab{b}})Wang, Chen, Ma, Zhou, Huang, Wang, Yang, He, Yu, Yang, et~al.]{wang2023lavie}
Yaohui Wang, Xinyuan Chen, Xin Ma, Shangchen Zhou, Ziqi Huang, Yi Wang, Ceyuan Yang, Yinan He, Jiashuo Yu, Peiqing Yang, et~al.
\newblock Lavie: High-quality video generation with cascaded latent diffusion models.
\newblock \emph{arXiv preprint arXiv:2309.15103}, 2023{\natexlab{b}}.

\bibitem[Wei et~al.(2023)Wei, Wang, Schuurmans, Bosma, Ichter, Xia, Chi, Le, and Zhou]{cot}
Jason Wei, Xuezhi Wang, Dale Schuurmans, Maarten Bosma, Brian Ichter, Fei Xia, Ed Chi, Quoc Le, and Denny Zhou.
\newblock Chain-of-thought prompting elicits reasoning in large language models, 2023.

\bibitem[Wermelinger(2023)]{wermelinger2023using}
Michel Wermelinger.
\newblock Using github copilot to solve simple programming problems.
\newblock In \emph{Proceedings of the 54th ACM Technical Symposium on Computer Science Education V. 1}, pages 172--178, 2023.

\bibitem[Xie et~al.(2022)Xie, Liu, and Fu]{xie2022gos}
Mingye Xie, Ting Liu, and Yuzhuo Fu.
\newblock Gos: A large-scale annotated outdoor scene synthetic dataset.
\newblock In \emph{ICASSP 2022-2022 IEEE International Conference on Acoustics, Speech and Signal Processing (ICASSP)}, pages 3244--3248. IEEE, 2022.

\bibitem[Xu et~al.(2022)Xu, Jiang, Wang, Fan, Shi, and Wang]{xu2022sinnerf}
Dejia Xu, Yifan Jiang, Peihao Wang, Zhiwen Fan, Humphrey Shi, and Zhangyang Wang.
\newblock Sinnerf: Training neural radiance fields on complex scenes from a single image.
\newblock In \emph{European Conference on Computer Vision}, pages 736--753. Springer, 2022.

\bibitem[Xu et~al.(2024)Xu, Ng, Wang, Sa, Duan, Li, Ji, and Li]{xu2024sketch2scene}
Yongzhi Xu, Yonhon Ng, Yifu Wang, Inkyu Sa, Yunfei Duan, Yang Li, Pan Ji, and Hongdong Li.
\newblock Sketch2scene: Automatic generation of interactive 3d game scenes from user's casual sketches.
\newblock \emph{arXiv preprint arXiv:2408.04567}, 2024.

\bibitem[Yang et~al.(2024{\natexlab{a}})Yang, Ding, Deng, Wang, and Qi]{yang2024regionplc}
Jihan Yang, Runyu Ding, Weipeng Deng, Zhe Wang, and Xiaojuan Qi.
\newblock Regionplc: Regional point-language contrastive learning for open-world 3d scene understanding.
\newblock In \emph{Proceedings of the IEEE/CVF Conference on Computer Vision and Pattern Recognition}, pages 19823--19832, 2024{\natexlab{a}}.

\bibitem[Yang et~al.(2024{\natexlab{b}})Yang, Jia, Zhi, and Huang]{yang2024physcene}
Yandan Yang, Baoxiong Jia, Peiyuan Zhi, and Siyuan Huang.
\newblock Physcene: Physically interactable 3d scene synthesis for embodied ai.
\newblock In \emph{Proceedings of Conference on Computer Vision and Pattern Recognition (CVPR)}, 2024{\natexlab{b}}.

\bibitem[Yu et~al.(2021)Yu, Ye, Tancik, and Kanazawa]{yu2021pixelnerf}
Alex Yu, Vickie Ye, Matthew Tancik, and Angjoo Kanazawa.
\newblock pixelnerf: Neural radiance fields from one or few images.
\newblock In \emph{Proceedings of the IEEE/CVF conference on computer vision and pattern recognition}, pages 4578--4587, 2021.

\bibitem[Zhang et~al.(2019)Zhang, Wang, Qin, Chen, and Gao]{zhang2019procedural}
Jian Zhang, Chang-bo Wang, Hong Qin, Yi Chen, and Yan Gao.
\newblock Procedural modeling of rivers from single image toward natural scene production.
\newblock \emph{The Visual Computer}, 35:\penalty0 223--237, 2019.

\bibitem[Zhang et~al.(2024)Zhang, Li, Wan, Wang, and Liao]{zhang2024text2nerf}
Jingbo Zhang, Xiaoyu Li, Ziyu Wan, Can Wang, and Jing Liao.
\newblock Text2nerf: Text-driven 3d scene generation with neural radiance fields.
\newblock \emph{IEEE Transactions on Visualization and Computer Graphics}, 2024.

\bibitem[Zhou et~al.(2024{\natexlab{a}})Zhou, Wang, Hou, Luo, Zhang, and Peng]{zhou2024scenex}
Mengqi Zhou, Yuxi Wang, Jun Hou, Chuanchen Luo, Zhaoxiang Zhang, and Junran Peng.
\newblock Scenex: Procedural controllable large-scale scene generation via large-language models.
\newblock \emph{arXiv preprint arXiv:2403.15698}, 2024{\natexlab{a}}.

\bibitem[Zhou et~al.(2024{\natexlab{b}})Zhou, Wang, Hou, Luo, Zhang, and Peng]{zhou2024scenexproceduralcontrollablelargescalescene}
Mengqi Zhou, Yuxi Wang, Jun Hou, Chuanchen Luo, Zhaoxiang Zhang, and Junran Peng.
\newblock Scenex:procedural controllable large-scale scene generation via large-language models, 2024{\natexlab{b}}.

\bibitem[Zhu et~al.(2024)Zhu, Chen, Ji, Ye, and Liu]{zhu2024llafs}
Lanyun Zhu, Tianrun Chen, Deyi Ji, Jieping Ye, and Jun Liu.
\newblock Llafs: When large language models meet few-shot segmentation.
\newblock In \emph{Proceedings of the IEEE/CVF Conference on Computer Vision and Pattern Recognition}, pages 3065--3075, 2024.

\bibitem[Zürn et~al.(2024)Zürn, Gladkov, Dudas, Cotter, Toteva, Shotton, Simaiaki, and Mohan]{zürn2024wayvescenes101datasetbenchmarknovel}
Jannik Zürn, Paul Gladkov, Sofía Dudas, Fergal Cotter, Sofi Toteva, Jamie Shotton, Vasiliki Simaiaki, and Nikhil Mohan.
\newblock Wayvescenes101: A dataset and benchmark for novel view synthesis in autonomous driving, 2024.

\end{thebibliography}
}
\clearpage
\setcounter{page}{1}
\maketitlesupplementary
\appendix
\section{Prompts Initialization}

\begin{figure*}[b]
    \centering

    \begin{subfigure}{0.45\textwidth}
        \centering
        \includegraphics[width=\linewidth]{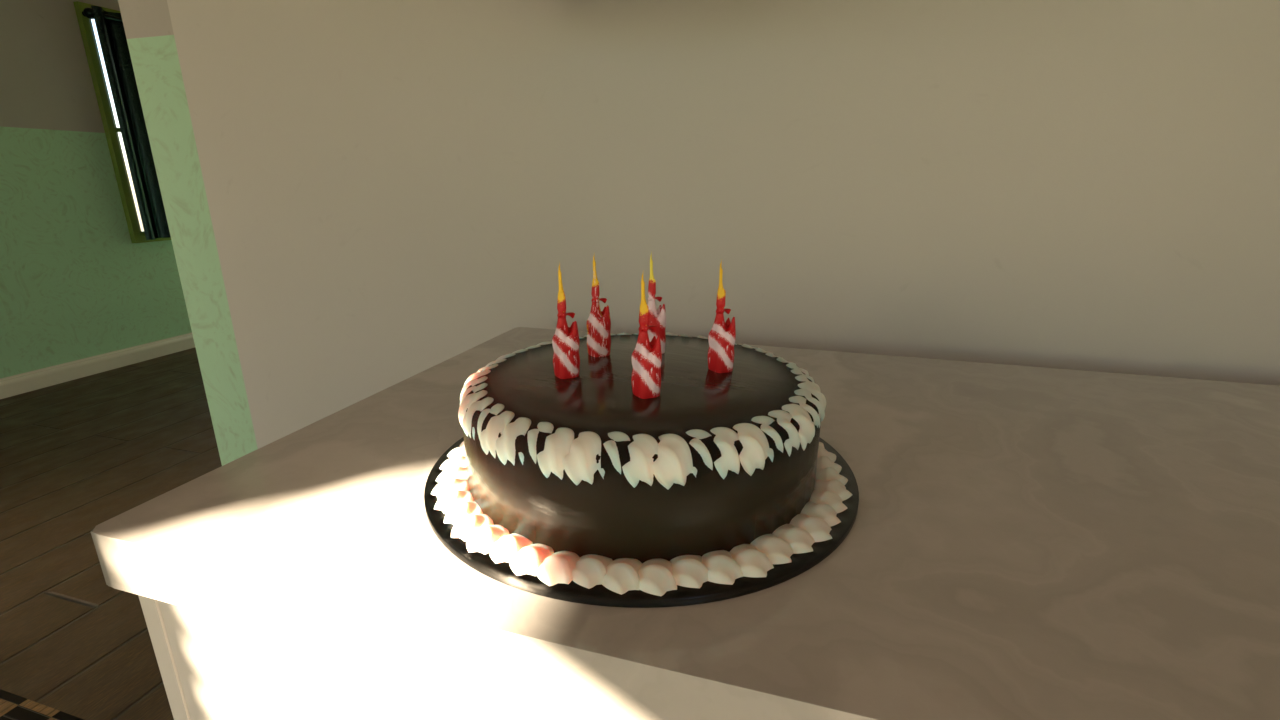}
        \caption{"a birthday cake in the plate"}
        \label{fig:birthday_cake}
    \end{subfigure}
    \begin{subfigure}{0.45\textwidth}
        \centering
        \includegraphics[width=\linewidth]{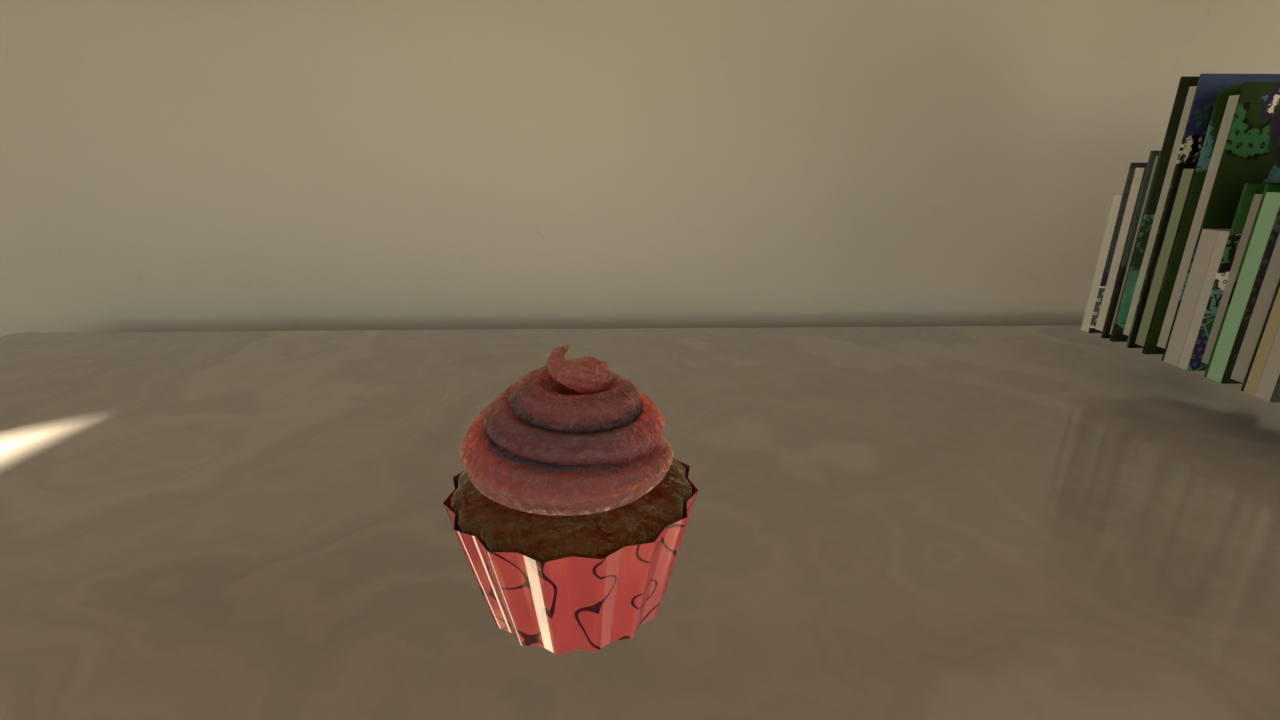}
        \caption{"video of a delicious dessert"}
        \label{fig:dessert}
    \end{subfigure}

    \vspace{0.1cm}

    \begin{subfigure}{0.45\textwidth}
        \centering
        \includegraphics[width=\linewidth]{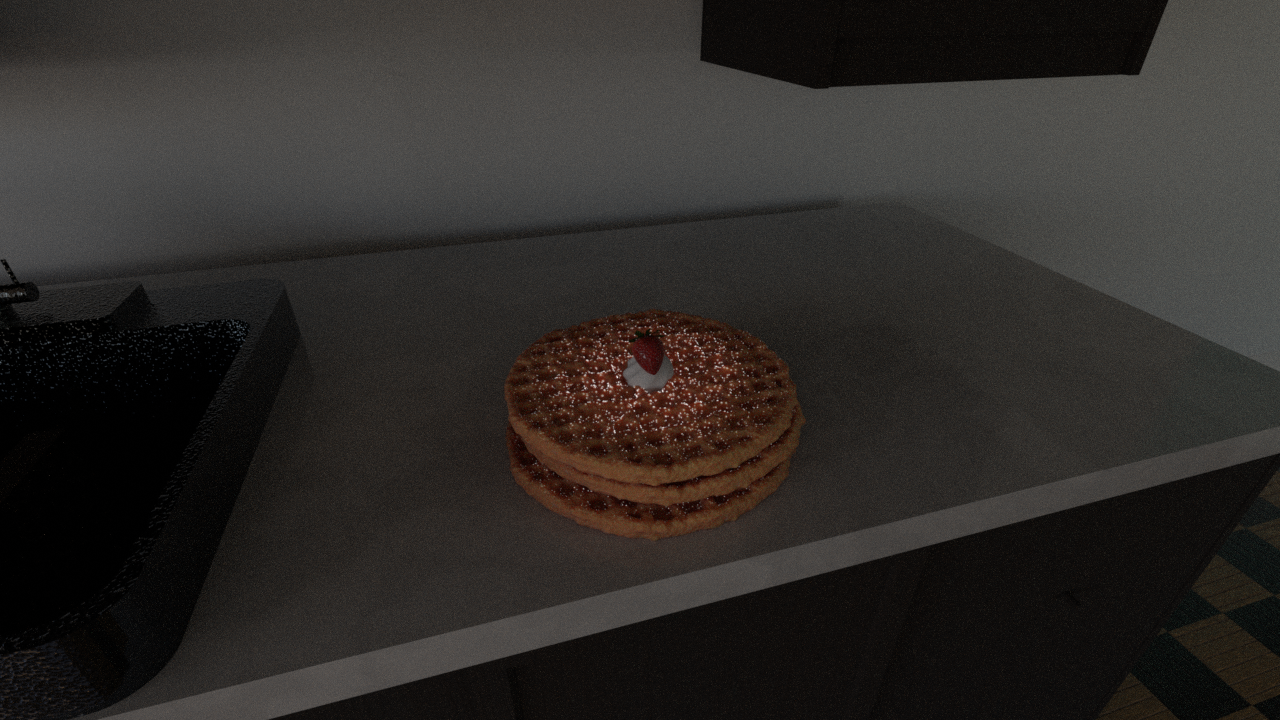}
        \caption{"waffles with whipped cream and fruit"}
        \label{fig:waffles}
    \end{subfigure}
    \begin{subfigure}{0.45\textwidth}
        \centering
        \includegraphics[width=\linewidth]{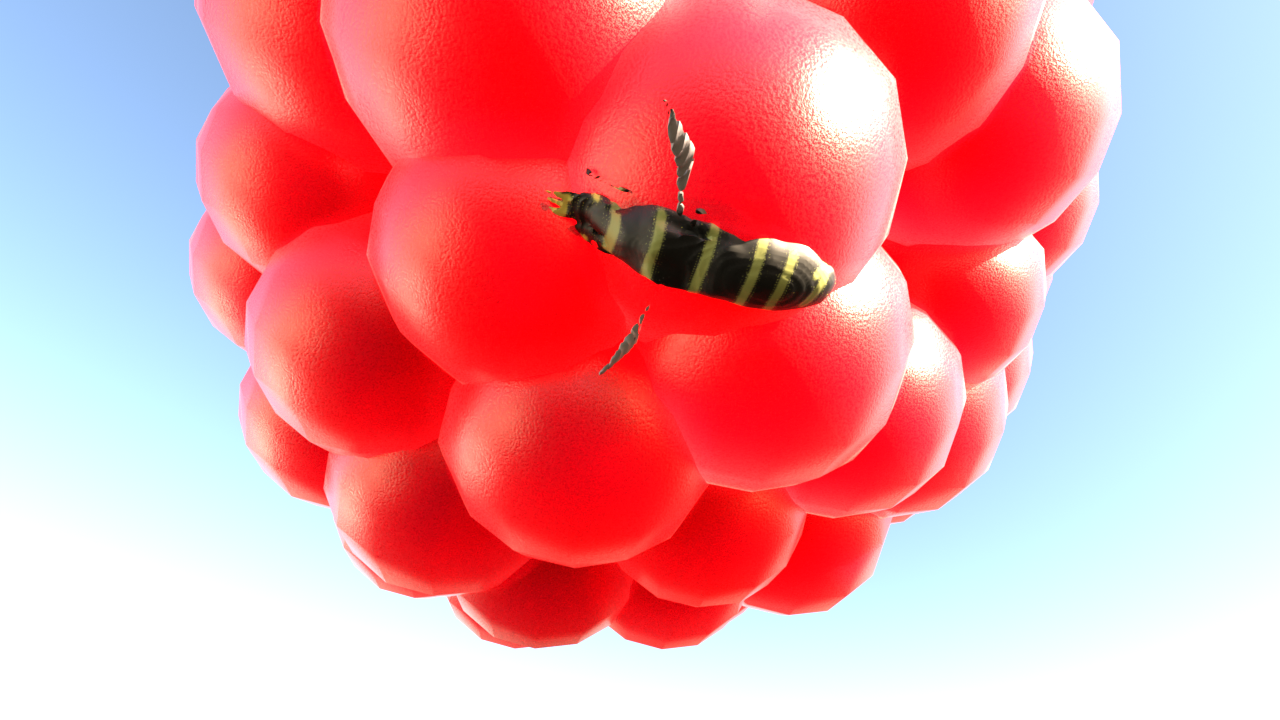}
        \caption{"focus shot of an insect at the bottom of a fruit"}
        \label{fig:insect}
    \end{subfigure}

    \caption{Screenshots from 4 different outputs from \pipeline{} and their respective prompts.}
    
    \label{fig:food}
    
\end{figure*}

\subsection{Initial Prompts For Scene Codex and BlenderGPT}
As presented in Table\,\ref{tab:codextable}, the initial system prompt for Scene Codex is provided. This table outlines the specific configuration and structure required for engaging with the Scene Codex framework, offering a clear overview of the input parameters that govern its functionality. Additionally, Table\,\ref{tab:blendergpttable} displays the initial prompt associated with BlenderGPT. We first prompt LLM to generate a guideline to accomplish the users' request through Chain-of-Thoughts. Then, another LLM generates detailed executable Python codes based on the provided steps.

\begin{table*}[h]
  \caption{Scene Codex System Prompt} 
      \noindent
      \small
      \label{tab:codextable} 
      \centering
      \scriptsize
      \begin{tabular}{p{1.1cm}|p{15cm}}
      & \makecell{The Prompts} \\
      \midrule
      Description: & You are an assistant for generating Infinigen code command based on natural language input. Use the following pieces of retrieved context to answer the question. If you don't know the answer, just say that you don't know. Here is the helping doc for python function manage\_jobs, you should only choose and include those options in your generated command after the starting code. You are an assistant for generating Infinigen code command based on natural language input. \
      Use the following pieces of retrieved context to answer the question. \
      If you don't know the answer, just say that you don't know. \
      Here is the helping doc for python function manage\_jobs, you should only choose and include those options in your generated command after the starting code. \\
      \midrule
      Usage: & manage\_jobs.py[-h] [-o OUTPUT\_FOLDER] [--num\_scenes NUM\_SCENES]\newline
                          [--meta\_seed META\_SEED]\newline
                          [--specific\_seed SPECIFIC\_SEED [SPECIFIC\_SEED ...]]\newline
                          [--use\_existing] [--warmup\_sec WARMUP\_SEC]\newline
                          [--cleanup (all,big\_files,none,except\_logs,except\_crashed)]\newline
                          [--configs [CONFIGS ...]] [-p OVERRIDES [OVERRIDES ...]]\newline
                          [--wandb\_mode (online,offline,disabled)]\newline
                          [--pipeline\_configs PIPELINE\_CONFIGS [PIPELINE\_CONFIGS ...]]\newline
                          [--pipeline\_overrides PIPELINE\_OVERRIDES [PIPELINE\_OVERRIDES ...]]\newline
                          [--overwrite] [-d] [-v] \\
      \midrule
      Options: & -h, --help Explanation:  show this help message and exit\newline
      -o OUTPUT\_FOLDER, --output\_folder OUTPUT\_FOLDER\newline
      --num\_scenes NUM\_SCENES. Explanation: Number of scenes to attempt before terminating\newline
      --meta\_seed META\_SEED. Explanation: What seed should be used to determine the random seeds of each scene? Leave as None unless deliberately replicating past runs\newline
      --specific\_seed SPECIFIC\_SEED [SPECIFIC\_SEED ...]. Explanation: The default, None, will choose a random seed per scene. Otherwise, all scenes will have the specified seed. Interpreted as an integer if possible.\newline
      --use\_existing  Explanation: If set, then assume output\_folder is an existing folder from a terminated run, and make a best-possible-effort to resume from where it left off\newline
      --warmup\_sec WARMUP\_SEC Explanation: Perform a staggered start over the specified period, so that jobs don't sync up or all write to disk at similar times.\newline
      --cleanup (all,big\_files,none,except\_logs,except\_crashed) Explanation: What files should be cleaned up by the manager as it runs\newline
      --configs [CONFIGS ...] Explanation: List of gin config names to pass through to all underlying scene generation jobs. Available gin files are: ['canyon.gin', 'plain.gin', 'under\_water.gin', 'fast\_terrain\_assets.gin', 'mountain.gin', 'kelp\_forest.gin', 'stereo\_training.gin', 'snowy\_mountain.gin', 'high\_quality\_terrain.gin', 'simple.gin', 'no\_creatures.gin', 'no\_rocks.gin', 'natural.gin', 'reuse\_terrain\_assets.gin', 'no\_assets.gin', 'dev.gin', 'tilted\_river.gin', 'arctic.gin', 'no\_particles.gin', 'simulated\_river.gin', 'coast.gin', 'use\_cached\_fire.gin', 'forest.gin', 'coral\_reef.gin', 'cliff.gin', 'snake.gin', 'use\_on\_the\_fly\_fire.gin', 'tiger.gin', 'base\_surface\_registry.gin', 'experimental.gin', 'river.gin', 'cave.gin', 'desert.gin']. You should primarily choose files from this list. Any new ",gin" file should have detailed additional config setting.\newline
      -p OVERRIDES [OVERRIDES ...], --overrides OVERRIDES [OVERRIDES ...] Explanation: List of gin overrides to pass through to all underlying scene generation jobs\newline
      --wandb\_mode (online,offline,disabled) Explanation: Mode kwarg for wandb.init(). Set up wandb before use.\newline
      --pipeline\_configs PIPELINE\_CONFIGS [PIPELINE\_CONFIGS ...] Explanation: List of gin config names from datagen/pipeline\_configs to configure this execution. Available gin files are: ['cuda\_terrain.gin', 'indoor\_background\_configs.gin', 'export.gin', 'base.gin', 'asset\_demo.gin', 'upload.gin', 'opengl\_gt.gin', 'blender\_gt.gin', 'gt\_test.gin', 'opengl\_gt\_noshortrender.gin', 'stereo.gin', 'block\_terrain\_experiment.gin', 'stereo\_1h\_jobs.gin', 'stereo\_video.gin', 'monocular\_video.gin', 'monocular\_flow.gin', 'monocular.gin', 'slurm\_cpuheavy.gin', 'slurm.gin', 'local\_256GB.gin', 'local\_64GB.gin', 'local\_128GB.gin', 'slurm\_high\_memory.gin', 'slurm\_1h.gin', 'local\_16GB.gin']\newline
      --pipeline\_overrides PIPELINE\_OVERRIDES [PIPELINE\_OVERRIDES ...] Explanation: List of gin overrides to configure this execution. Use command `--pipeline\_overrides` rather than `--overrides` when a gin file override the `manage\_jobs.py` process, not the main `Infinigen\_examples/generate\_nature.py` driver.\newline
      --overwrite Explanation: Overwrite the output folder if it already exists.\newline
      -d, --debug
      -v, --verbose \\
      \midrule
      Description: & Keep the answer concise. You should always generate Infinigen commands in a single line starting with "python -m Infinigen.datagen.manage\_jobs". You can provide options to user if you need based on retrieved context. And you can write your own ".gin" file from the retrieved context if needed. You can propose to change seed/including simple scene setting gin files due to insufficient GPU memory or crashed animation with max\_full\_retries=30 and max\_step\_tries=25 
      You should only include parameters in the context, do not hallucinate or improvise any non-existent parameters because of the request. Say you don't know how to do it if you don't know.
      
      \end{tabular}
      \vspace*{-0.2cm}
\end{table*}

%


\begin{table*}[h]
  \caption{BlenderGPT Prompt} 
      \noindent
      \small
      \label{tab:blendergpttable} 
      \centering
      \scriptsize
      \begin{tabular}{p{1.1cm}|p{15cm}}
      & \makecell{The Prompts} \\
      \midrule
    Description & You are an assistant made for the purpose of helping the user with Blender, the 3D software. \newline
- Respond with your answers in markdown (```).\newline 
- Preferably import entire modules instead of bits.\newline
- Do not perform destructive operations on the meshes.\newline
- Do not use cap\_ends. Do not do more than what is asked (setting up render settings, adding cameras, etc)\newline
- Do not respond with anything that is not Python code.\newline
- There is a pre-generated scene.blend with objects, animations, and cameras in the scene. You should primarily use those objects in the scene and generate related code.\newline
- You will be given steps from another Chain-of-Thought agent to help accomplish the user-requested task. Generate python codes following the directions. \\

\midrule
CoT Prompt: & Follow these steps: \newline
1. Think through the problem step by step within the \textlangle thinking\textrangle tags. \newline
2. Reflect on your thinking to check for any errors or improvements within the \textlangle reflection\textrangle tags.\newline
3. Make any necessary adjustments based on your reflection. \newline
4. Provide your final, concise answer within the \textlangle output\textrangle tags. \newline
Important: The \textlangle thinking\textrangle and \textlangle reflection\textrangle sections are for your internal reasoning process only. \newline
Do not include any part of the final answer in these sections. \newline
The potential related object names should be gathered from those extra context. \newline
The actual response to the query must be entirely contained within the \textlangle output\textrangle tags. \\

\midrule
Format: & Use the following format for your response: \newline
\textlangle thinking\textrangle\newline
[Your step-by-step reasoning goes here. This is your internal thought process, not the final answer.]\newline
\textlangle reflection\textrangle\newline
[Your reflection on your reasoning, checking for errors or improvements]\newline
\textlangle /reflection\textrangle\newline
[Any adjustments to your thinking based on your reflection]\newline
\textlangle /thinking\textrangle\newline
\textlangle output\textrangle\newline
[Your final, concise answer to the query. This is the only part that will be shown to the next agent.]\newline
\textlangle /output\textrangle\newline
\end{tabular}
\vspace*{-0.2cm}
\end{table*}

\subsection{Few-shot examples}
As presented in Table\,\ref{tab:fewshotexample}, several \emph{description-command} pairs, used as few-shot examples for generating scenes with Infinigen, are provided. They are used as additional guidelines for Scene Codex to generate faithful, correct Infinigen commands.


\begin{table*}[h]
  \caption{Few-Shot Examples} 
      \noindent
      \small
      \label{tab:fewshotexample} 
      \centering
      \scriptsize
      \begin{tabular}{p{0.2cm}|p{15cm}}
      & \makecell{The Examples} \\
      \midrule

1 & "Description": "Generate an Infinigen command that will create a low quality desert scene", \par
"Command": "python -m Infinigen.datagen.manage\_jobs output\_folder outputs/hello\_world --num\_scenes 1 --specific\_seed 0 --configs desert.gin simple.gin --pipeline\_configs local\_16GB.gin monocular.gin blender\_gt.gin --pipeline\_overrides LocalScheduleHandler.use\_gpu=False" \\

\midrule
2 & "Description": "Generate an Infinigen command that will create 10000 large-scale high-quality stereo scenes", \par
"Command": "python -m Infinigen.datagen.manage\_jobs --output\_folder outputs/stereo\_data --num\_scenes 10000 --pipeline\_configs slurm.gin stereo.gin cuda\_terrain.gin --cleanup big\_files --warmup\_sec 60000 --config high\_quality\_terrain" \\

\midrule
3 & "Description": "Generate an Infinigen command that will create high quality 500 videos", \par
"Command": "python -m Infinigen.datagen.manage\_jobs --output\_folder outputs/my\_videos --num\_scenes 500 --pipeline\_configs slurm.gin monocular\_video.gin cuda\_terrain.gin --cleanup big\_files --warmup\_sec 60000 --config trailer\_video high\_quality\_terrain" \\

\midrule
4 & "Description": "Generate an Infinigen command that will create a few (50) low-resolution scenes", \par
"Command": "python -m Infinigen.datagen.manage\_jobs --output\_folder outputs/dev --num\_scenes 50 --pipeline\_configs slurm.gin monocular.gin cuda\_terrain.gin --cleanup big\_files --warmup\_sec 1200 --configs dev" \\

\midrule
5 & "Description": "Generate an Infinigen command that will create a image that always have rain:", \par
"Command": "python -m Infinigen.datagen.manage\_jobs --output\_folder outputs/my\_videos --num\_scenes 500 --pipeline\_configs slurm.gin monocular.gin cuda\_terrain.gin --cleanup big\_files --warmup\_sec 30000  --overrides compose\_nature.rain\_particles\_chance=1.0" \\

\midrule
6 & "Description": "Generate an Infinigen command that will create a high quality arctic scene with windy weather and output a 20s video", \par
"Command": "python -m Infinigen.datagen.manage\_jobs --num\_scenes 1 --pipeline\_configs monocular\_video.gin cuda\_terrain.gin local\_64GB.gin --configs arctic.gin high\_quality\_terrain --pipeline\_overrides iterate\_scene\_tasks.frame\_range=[1,480] --overrides compose\_nature.wind\_chance=1.0 --output\_folder outputs/windy\_arctic'" \\

\end{tabular}
\vspace*{-0.2cm}
\end{table*}

\section{Detailed Procedure}
The detailed procedure for \pipeline{} is presented in Alg.\ref{alg:pseudo-code}. In this algorithm, we include additional details and processes of our proposed framework. The pseudo-code provides a comprehensive overview that accurately reflects the core functionalities and aids the understanding of the workflow and operational mechanics shown in Figure\,\ref{fig:pipeline}. Each step is carefully articulated to ensure that the logical flow and interactions between components are clearly conveyed.
%


\begin{algorithm*}
\caption{Generate and Enhance Code with Human Feedback}
\label{alg:pseudo-code}
\KwIn{$\textit{user\_input}$ \text{(text input from the user, describing the task)}}
\KwOut{$\textit{final\_video}$ \text{(the final rendered video after code enhancement)}}

\hrule

\textbf{Step 1:} Receive the user's input \\
$\textit{user\_input} \gets \textit{get\_input()}$ \tcp{The user provides a textual prompt, which will be used for code generation and scene creation} 

\textbf{Step 2:} Generate initial code based on user input using the Scene Codex with RAG (Retrieval-Augmented Generation) \\
$\textit{init\_code} \gets \textit{Codex}(\textit{user\_input}, \textit{RAG})$ \tcp{Scene Codex generates code based on the user's prompt, leveraging RAG for relevant data retrieval} 

\textbf{Step 3:} Generate the initial coarse scene from Infinigen with the initial command \\
$\textit{coarse\_scene} \gets \textit{Infinigen}(\textit{init\_code})$ \tcp{Infinigen creates the coarse scene from generated initial command} 

\textbf{Step 4:} Format the generated scene for readability and structure using LLMs (Large Language Models) \\
$\textit{formatted\_scene} \gets \textit{format}(\textit{coarse\_scene}, \textit{LLMs})$ \tcp{LLMs are used to refine and structure the generated scene for better clarity and usability} 

\textbf{Step 5:} Store the formatted scene in the coarse-specific code repository \\
$\textit{coarse\_RAG.append}(\textit{USDA\_file(formatted\_scene)})$ \tcp{The formatted scene is converted to coarse RAG database for future reference and use} 

\textbf{Step 6:} Edit the 3D scene based on the user input using BlenderGPT with multiple tools \\
$\textit{edited\_scene} \gets \textit{BlenderGPT}(\textit{user\_input}, \textit{Procedural Dataset}, \textit{Shap-e}, \textit{coarse\_RAG},\textit{Human})$ \tcp{BlenderGPT revise and edit the 3D scene by combining the user input with LLM, procedural dataset, Shap-e, and coarse\_RAG} 


\textbf{Step 7:} Update the Infinigen scene based on the edited\_scene from step 6 \\
$\textit{fine\_scene} \gets \textit{Infinigen}(\textit{edited\_scene})$ \tcp{Infinigen updates the scene from the edited scene and outputs a fine-detailed scene} 


\textbf{Step 8:} Format the improved code for readability and structure using LLMs \\
$\textit{imp\_formatted\_code} \gets \textit{format}(\textit{fine\_scene}, \textit{LLMs})$ \tcp{The improved scene is formatted using LLMs to ensure it is well-structured and clear} 

\textbf{Step 9:} Store the improved code in the refined code repository for further processing \\
$\textit{refined\_RAG.append}(\textit{USDA\_file(imp\_formatted\_code)})$ \tcp{The improved code is added to the refined RAG, which holds optimized code for the project} 

\textbf{Step 10:} Render the final video using Infinigen, combining the coarse and refined code repositories with BlenderGPT and Human \\
$\textit{final\_video} \gets \textit{Infinigen}(\textit{user\_input}, \textit{BlenderGPT}, \textit{Procedural Dataset},\textit{Shap-E}, \textit{coarse\_RAG}, \textit{refined\_RAG},\textit{Human})$ \tcp{Infinigen creates the final video by merging the coarse RAG and the refined RAG, using BlenderGPT and Procedural Dataset for 3D rendering} 

\textbf{Step 11:} Return the final rendered video to the user \\
\KwRet{$\textit{final\_video}$} \tcp{The final video, which has been generated and enhanced based on user input and feedback, is returned to the user} 
\end{algorithm*}

\section{Additional Video Examples}
We further rendered additional video examples to demonstrate our long-video generation capabilities. We created one 10-minute video, two 2-minute videos,  three 1-minute videos, and three 30-second videos. We demonstrated Human in the Loop with BlenderGPT in a short video. 

We have also included more examples of outputs from \pipeline{} in Figure \ref{fig:food}. We have included these examples to showcase the utility of our procedural dataset and the capabilities of \pipeline{} beyond the creation of generic environments and scenery. Figure \ref{fig:food} focuses on prompts that relate to a subject or focal point rather than the description of an environment/setting, relying on our procedural dataset to create these subjects in each scene.

\section{Video Samples}
\begin{figure*}[h]
  \centering
  \includegraphics[width=1\linewidth]{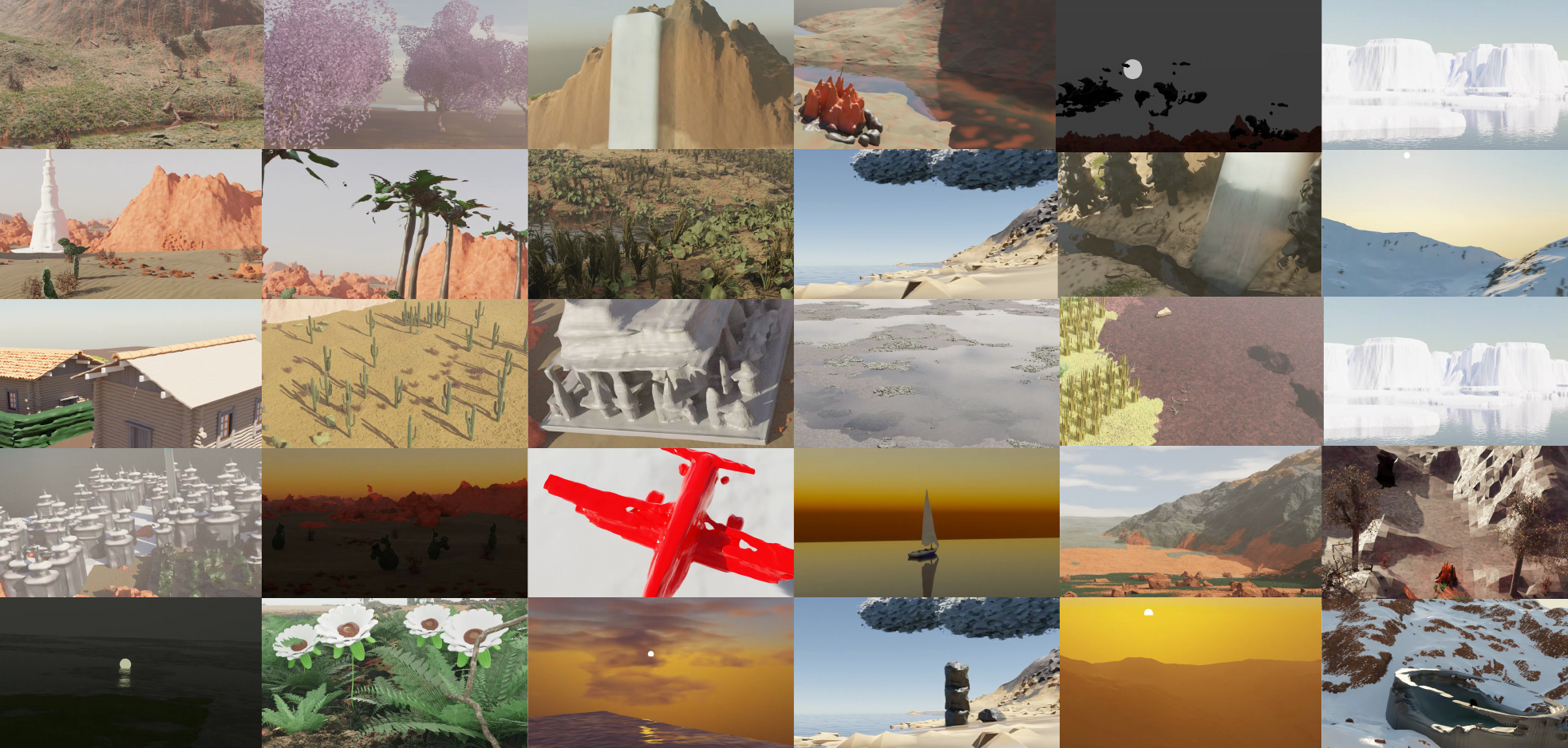}
  \caption{Thirty samples of videos generated by the framework are presented, showcasing the system's capability to produce diverse and high-quality outputs. Each video is represented by a key frame from the rendered video. }
  \label{fig:videosample}
\end{figure*}

We include more video samples produced by the \pipeline{} in Figure\,\ref{fig:videosample}. We show a single key frame of each video in various scenarios, showcasing \pipeline{}'s ability to handle diverse inputs and produce coherent, contextually relevant outputs. This figure serves as a visual representation of our framework's potential, providing a clear indication of its applicability to real-world tasks.

\section{New Assets}

\begin{figure*}[h]
  \centering
  \includegraphics[width=1\linewidth]{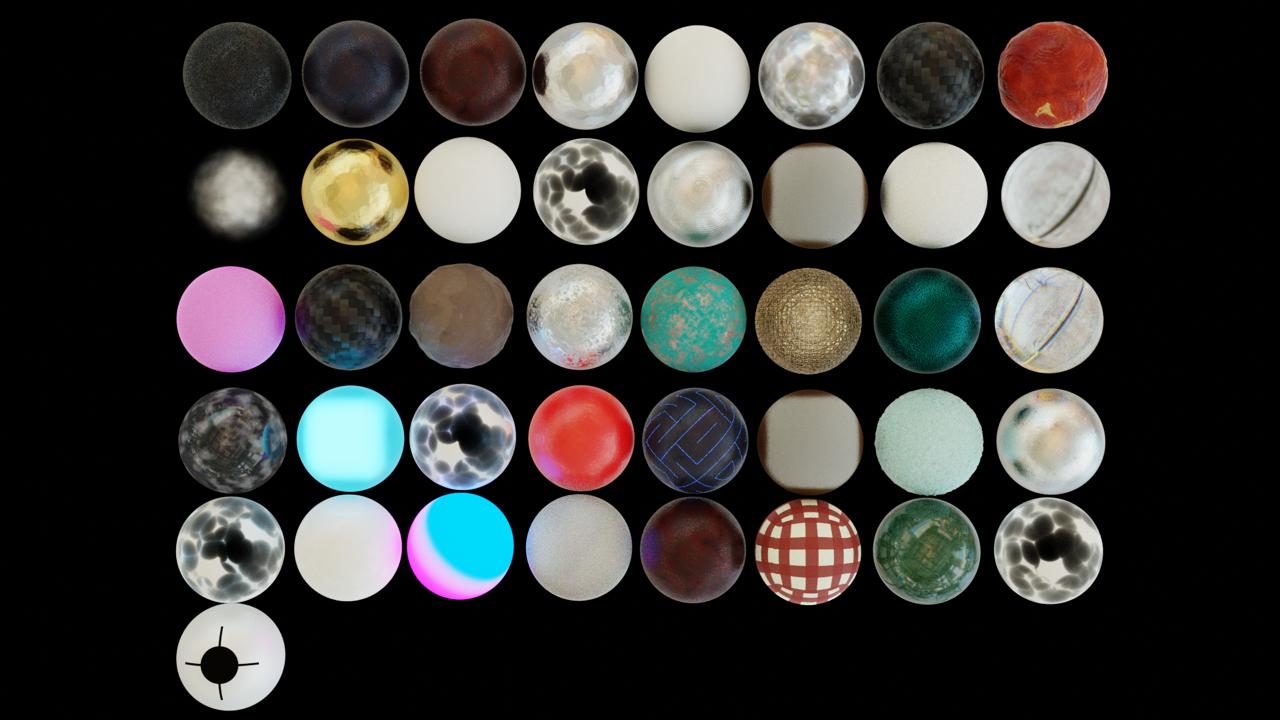}
  \caption{The dataset has been updated with 41 new materials. This addition enhances the diversity and scope of the dataset, enabling it to better accommodate a wider range of inputs and user requirements. }
  \label{fig:new-materials}
\end{figure*}

\begin{figure*}[tb]
    \centering

    \begin{subfigure}{0.45\textwidth}
        \centering
        \includegraphics[width=\linewidth]{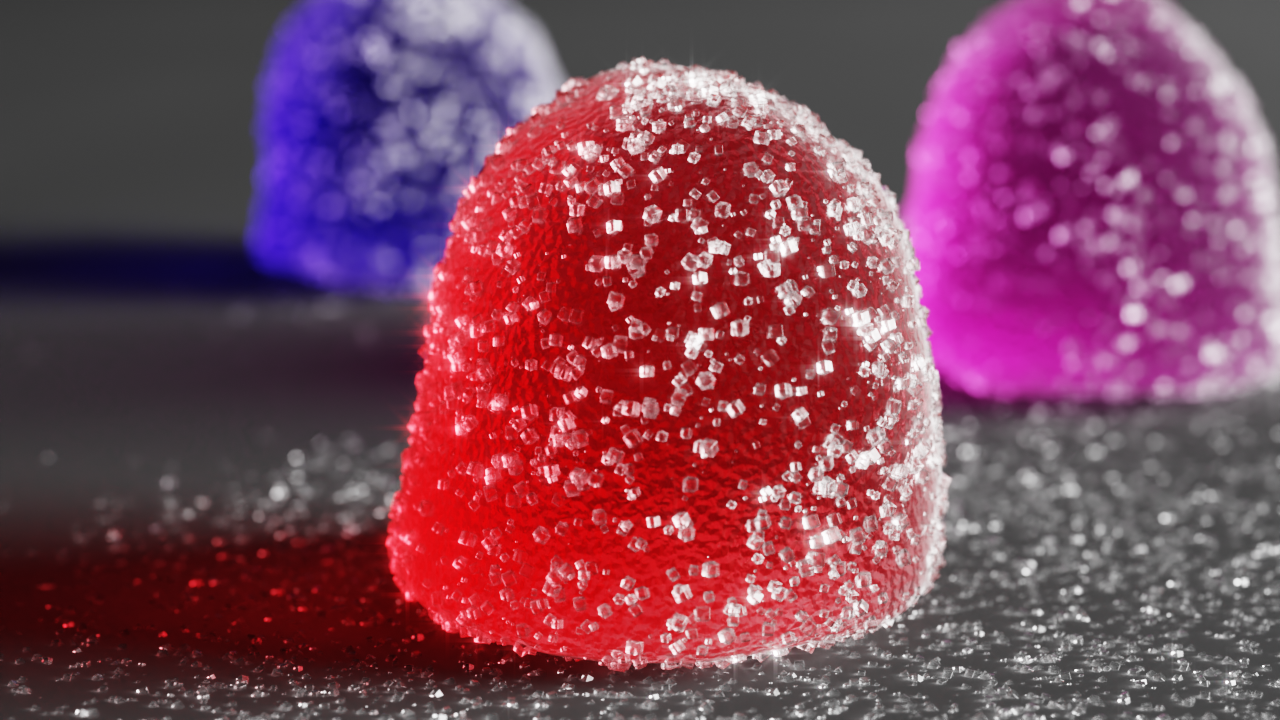}
        \caption{Sugar coated candy}
        \label{fig:sugar_coated_candy}
    \end{subfigure}
    \begin{subfigure}{0.45\textwidth}
        \centering
        \includegraphics[width=\linewidth]{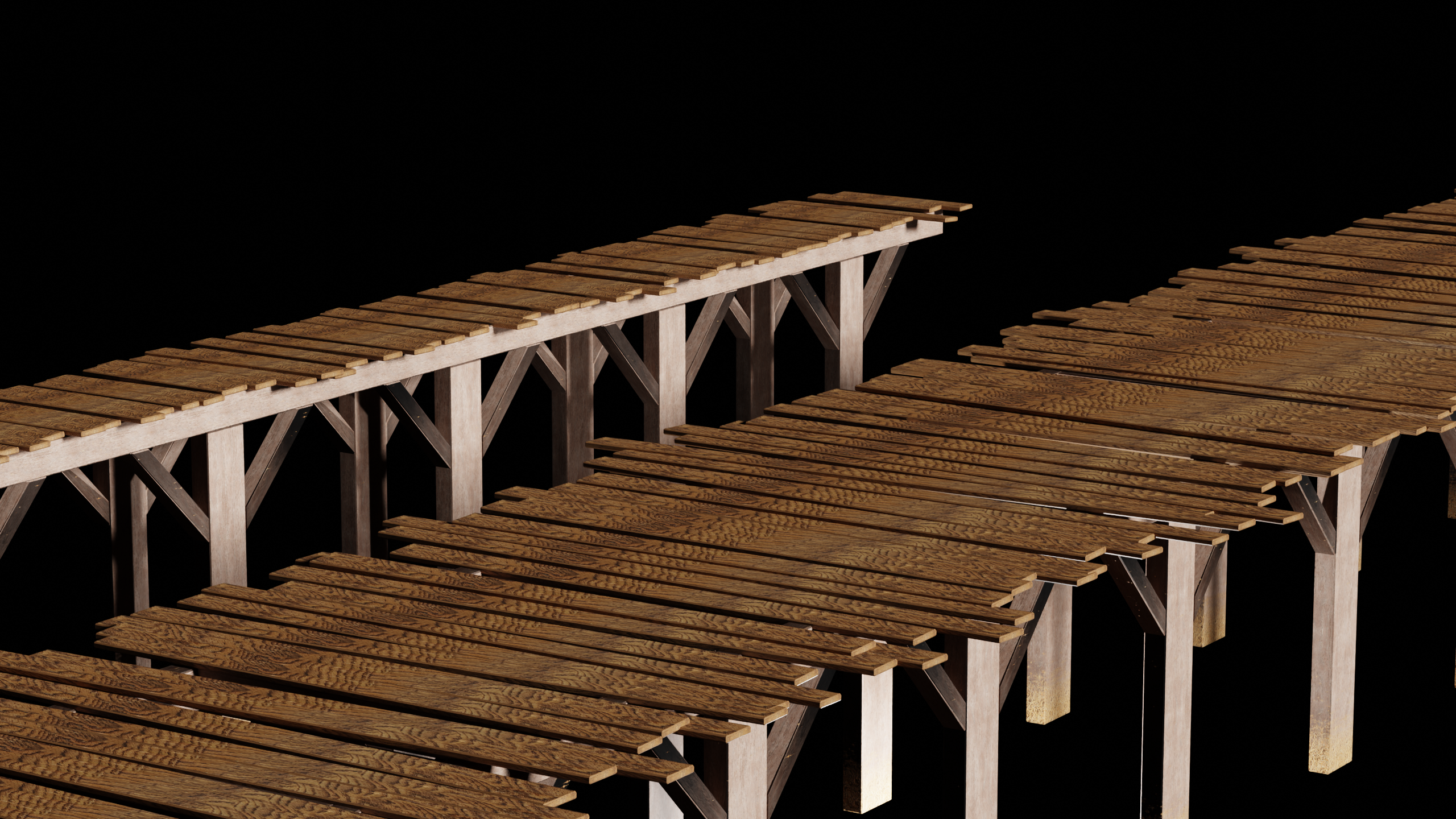}
        \caption{Procedural bridge generator}
        \label{fig:bridge}
    \end{subfigure}

    \vspace{0.1cm}

    \begin{subfigure}{0.45\textwidth}
        \centering
        \includegraphics[width=\linewidth]{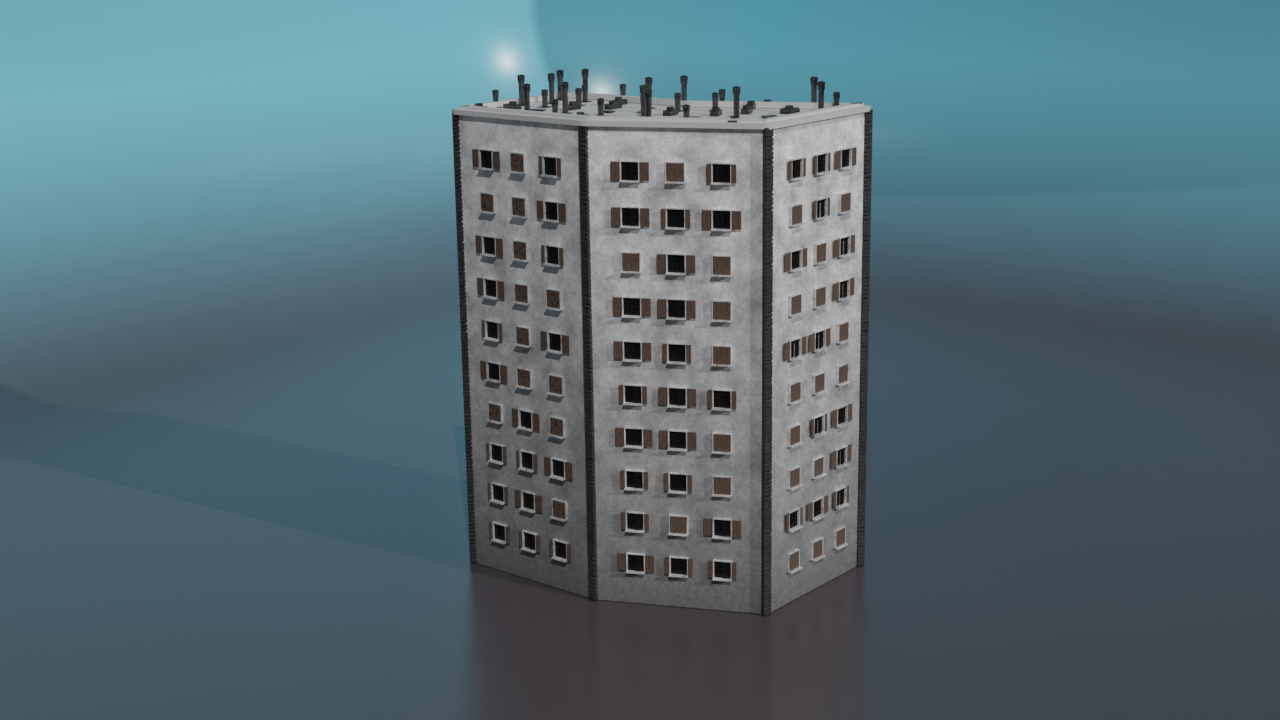}
        \caption{Apartment building}
        \label{fig:building}
    \end{subfigure}
    \begin{subfigure}{0.45\textwidth}
        \centering
        \includegraphics[width=\linewidth]{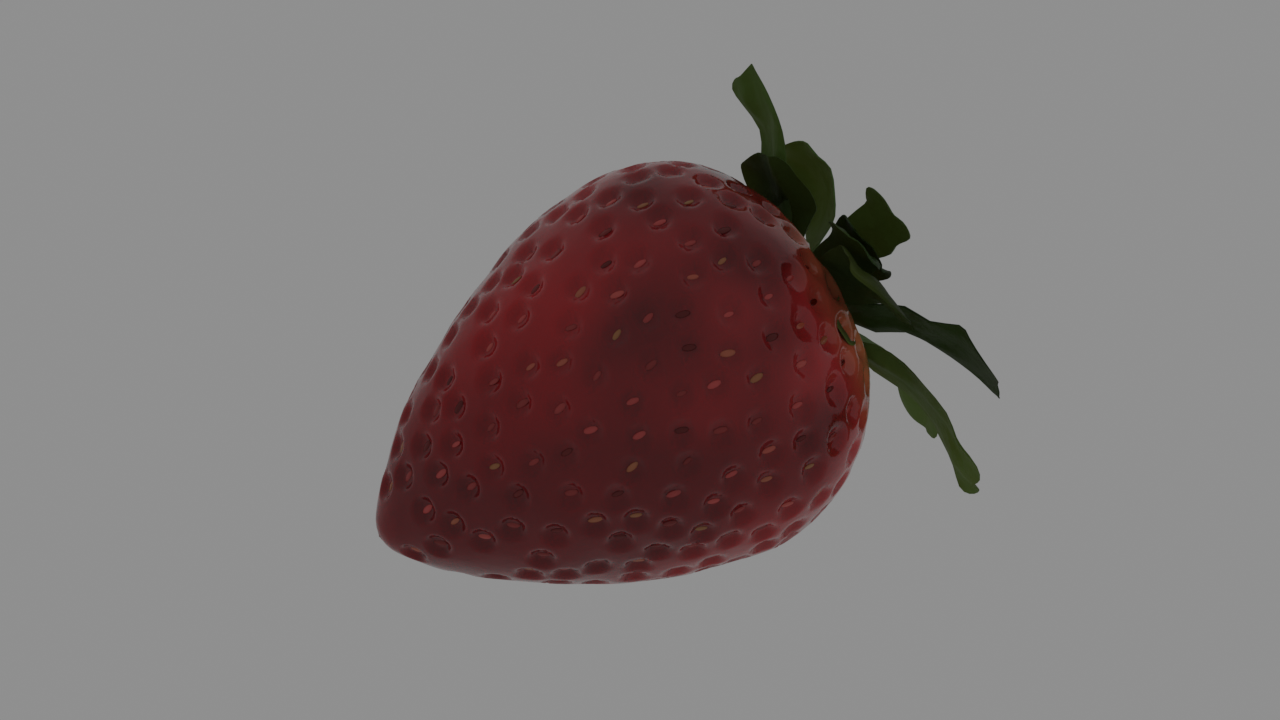}
        \caption{Strawberry with customizable geometry nodes}
        \label{fig:strawberry}
    \end{subfigure}

    \caption{Four of the new assets added to the dataset from the internet. Each image was rendered inside of its original blender scene. Note: (d) was used in Fig. \ref{fig:waffles}, demonstrating the expansion of the dataset's scope and consequently \pipeline{}'s capabilities as new objects are integrated into the dataset.}
    
    \label{fig:new-objects}
    
\end{figure*}

We have expanded our procedural database by adding an additional 41 material nodes assets and 12 geometry nodes assets to demonstrate the ability to extend the dataset for more precise alignment with user prompts, as shown in Figure\,\ref{fig:new-materials} and Figure \ref{fig:new-objects}. This enhancement improves the system’s flexibility and ability to generate more tailored outputs for a wider range of user queries.

\end{document}